\documentclass[sigconf]{acmart}
\usepackage{booktabs, colortbl}
\usepackage{epstopdf}
\usepackage{multirow}

\AtBeginDocument{%
  \providecommand\BibTeX{{%
    \normalfont B\kern-0.5em{\scshape i\kern-0.25em b}\kern-0.8em\TeX}}}

\renewcommand\footnotetextcopyrightpermission[1]{}
\setcopyright{none}
\begin{document}

\title{AesExpert: Towards Multi-modality Foundation Model for \\ Image Aesthetics Perception }


\author{Yipo Huang }
\authornote{This research was done during his visit to Nanyang Technological University in Singapore under the supervision of Prof. Weisi Lin.}
\affiliation{%
  \institution{School of Artificial Intelligence}
  \city{Xidian University}
   \country{}}
\email{huangyipo@stu.xidian.edu.cn}

\author{Xiangfei Sheng}
\affiliation{%
  \institution{School of Artificial Intelligence}
  \city{Xidian University}
     \country{}}
\email{xiangfeisheng@gmail.com}

\author{Zhichao Yang}
\affiliation{%
  \institution{School of Artificial Intelligence}
  \city{Xidian University}
    \country{}}
\email{yangzhichao@stu.xidian.edu.cn}

\author{Quan Yuan}
\affiliation{%
  \institution{School of Artificial Intelligence}
  \city{Xidian University}
    \country{}}
\email{dylan.yuanquan@stu.xidian.edu.cn}

\author{Zhichao Duan}
\affiliation{%
  \institution{School of Artificial Intelligence}
  \city{Xidian University}
    \country{}}
\email{zach@stu.xidian.edu.cn}

\author{Pengfei Chen}
\affiliation{%
  \institution{School of Artificial Intelligence}
  \city{Xidian University}
    \country{}}
\email{chenpengfei@xidian.edu.cn}

\author{Leida Li}
\authornote{Corresponding author.}
\affiliation{%
  \institution{School of Artificial Intelligence}
  \city{Xidian University}
    \country{}}
\email{ldli@xidian.edu.cn}

\author{Weisi Lin}
\affiliation{%
  \institution{College of Comput. and Data Science}
  \city{Nanyang Technological University}
    \country{}}
\email{wslin@ntu.edu.sg}

\author{Guangming Shi}
\affiliation{%
  \institution{School of Artificial Intelligence}
  \city{Xidian University}
    \country{}}
\email{gmshi@xidian.edu.cn}
\renewcommand{\shortauthors}{Yipo Huang and Xiangfei Sheng, et al.}

\begin{abstract}
  The highly abstract nature of image aesthetics perception (IAP) poses significant challenge for current multimodal large language models (MLLMs). The lack of human-annotated multi-modality aesthetic data further exacerbates this dilemma, resulting in MLLMs falling short of aesthetics perception capabilities. To address the above challenge, we first introduce a comprehensively annotated \textbf{Aes}thetic \textbf{M}ulti-\textbf{M}odality \textbf{I}nstruction \textbf{T}uning (\textbf{AesMMIT}) dataset, which serves as the footstone for building multi-modality aesthetics foundation models. Specifically, to align MLLMs with human aesthetics perception, we construct a corpus-rich aesthetic critique database with 21,904 diverse-sourced images and 88K human natural language feedbacks, which are collected via progressive questions, ranging from coarse-grained aesthetic grades to fine-grained aesthetic descriptions. To ensure that MLLMs can handle diverse queries, we further prompt GPT to refine the aesthetic critiques and assemble the large-scale aesthetic instruction tuning dataset, \textit{i.e.} \textbf{AesMMIT}, which consists of 409K multi-typed instructions to activate stronger aesthetic capabilities. Based on the AesMMIT database, we fine-tune the open-sourced general foundation models, achieving multi-modality \textbf{Aes}thetic \textbf{Expert} models, dubbed \textbf{AesExpert}. Extensive experiments demonstrate that the proposed AesExpert models deliver significantly better aesthetic perception performances than the state-of-the-art MLLMs, including the most advanced GPT-4V and Gemini-Pro-Vision. Project homepage: \url{https://yipoh.github.io/aes-expert/}.
  \end{abstract}

  


\begin{teaserfigure}
  \includegraphics[width=\textwidth]{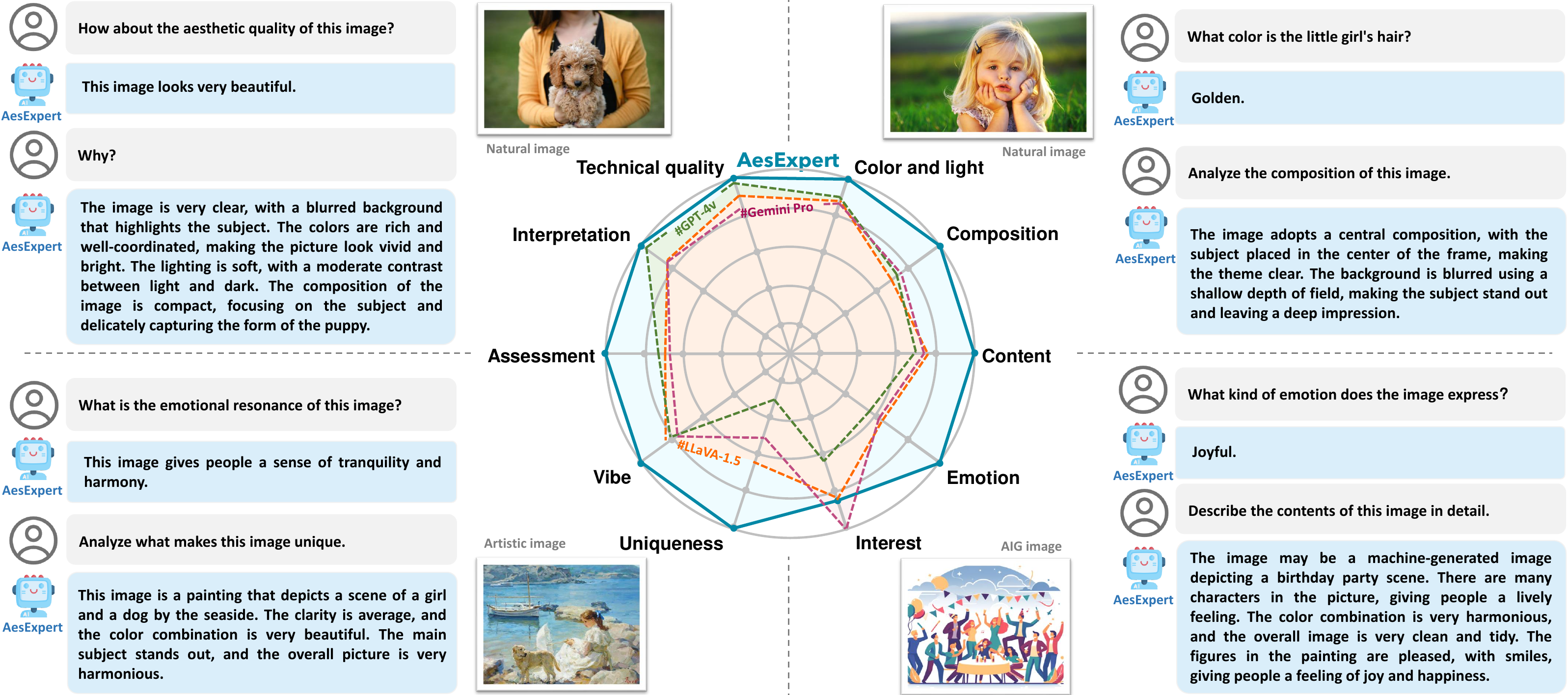}
  \caption{Performance of the proposed \textbf{AesExpert} on various aesthetic perception dimensions, in comparison with the most advanced GPT-4V and Gemini-Pro-Vision, as well as the open-sourced LLaVA-1.5-13B.}  
  \label{teaser}
\end{teaserfigure}


\maketitle

\thispagestyle{empty}
\pagestyle{empty}

\section{Introduction}
Multimodal large language models (MLLMs) have attracted significant attention in the research community. These foundation models, like GPT-4V \cite{GPT-4V} and LLaVA \cite{LLAVA}, have demonstrated remarkable progress in serving as general-purpose visual assistants, capable of interacting and collaborating with users \cite{Q-Align}. Concurrently, their outstanding cross-modality capacities further promote a paradigm shift in the computer vision domain, with researchers seeking to transcend the limitations of traditional task-specific approaches and develop multi-modality foundation models that generalize effectively across various visual tasks \cite{mi2023clif, hui2023rate, IET}. Despite the advancements achieved, experiments on current MLLMs reveal obvious limitations in the \textbf{highly-abstract image aesthetics perception} task \cite{AesBench, Zhu2}, which covers not only the extensively-studied image aesthetics assessment (IAA) \cite{Zhu3, zhu2023personalized, TAVAR}, but also fine-grained aesthetic attribute evaluation (\textit{e.g.} \textit{color, light,} and \textit{composition}), aesthetic emotion analysis, and image aesthetics caption \cite{VILA, CLIPAes, SARQUE}. The aesthetic perception abilities of MLLMs are crucial for a wide range of practical applications, such as smart photography, album management, photo recommendation, and image enhancement \cite{DSS, hui2022spatio}. \textbf{Consequently, it is urgent to build a unified foundation model that possesses general abilities across these aesthetic tasks and can precisely respond to open-ended human queries on image aesthetics perception.}

Due to the inherently data-hungry nature, current MLLMs rely on vast amounts of multi-modality instruction-following data to meet general-purpose visual and language understanding \cite{Qwen-VL}. Therefore, researchers have constructed numerous instruction fine-tuning datasets, such as COCO-VQA \cite{VQA}, Science QA \cite{ScienceQA} and LLaVA-Instruct-150K \cite{LLaVA-1.5}. However, existing instruction fine-tuning datasets are mainly engineered to enhance the general capacities of MLLMs, \emph{e.g.} visual question answering \cite{VQA}, image captioning \cite{Captioning}, object segmentation \cite{Segmentation} and content understanding \cite{mi2023vva, hui2022multi}. \textbf{A notable gap in these datasets is the inadequate focus on visual aesthetics.} To address the dilemma, we construct a comprehensively annotated \textbf{Aes}thetic \textbf{M}ulti-\textbf{M}odality \textbf{I}nstruction \textbf{T}uning (\textbf{AesMMIT}) dataset, based on which we further fine-tune the open-sourced general foundation models, achieving multi-modality \textbf{Aes}thetic \textbf{Expert} models, dubbed \textbf{AesExpert}, which delivers significantly better aesthetic perception performances than the state-of-the-art MLLMs. Intuitive comparisons and examples are shown in Figure \ref{teaser}. Specifically, this work includes three stages: 

\textit{Stage 1: Collecting human aesthetic feedback from subjective experiments.} To bridge the gap between MLLMs and human aesthetics perception, we invite human subjects to provide direct feedback on aesthetic perception and understanding via progressive questions, including three parts: 1) The \textbf{coarse-grained aesthetic evaluation} (\textit{e.g. This image looks quite beautiful/unattractive.}). 2) The \textbf{fine-grained reasoning and explanation} based on elemental aesthetic attributes (\textit{e.g. clarity, color, light, image object} and \textit{composition, etc.}). 3) The \textbf{finer-grained description} on aesthetic feeling (\textit{e.g. novel shooting view, interesting content} and \textit{expressed emotions}). With the three parts, the collected human feedbacks, denoted as \textbf{AesFeedback}, can capture the basic aesthetic perceptions and the evaluation reasoning process. The AesFeedback dataset contains \textbf{88K human critiques} on \textbf{21,904 multi-sourced images}.

\textit{Stage 2: Refining feedback with GPT for instruction-following data.} The constructed \textbf{AesFeedback} dataset plays a crucial role in fine-tuning MLLMs based on aesthetic instructions. However, to fully harness aesthetic capabilities, the dataset should also encompass an aesthetic question-answering component. To obtain rich question-answer pairs, inspired by the existing works \cite{ShareGPT4V}, we leverage GPT-4 to transform human feedback into instruction-following formats, which include both open-ended and multiple-choice question-answer pairs. To ensure that MLLMs can handle diverse queries, the instruction-following pairs cover diversified aesthetic perception dimensions (\textit{e.g. quality, attribute, emotion, interpretation, enhancement}, and \textit{context reasoning}), and commonly-used question types (\emph{e.g.} \emph{Yes-or-No, What, How, Why}, and other \emph{open-ended} questions). Through the above operations, we obtain the final \textbf{AesMMIT} dataset, which consists of \textbf{409K multi-typed instructions} to activate stronger aesthetic capabilities.

\textit{Stage 3: Building multi-modality aesthetics foundation model based on the proposed AesMMIT.} We introduce the instruction fine-tuning to improve the open-sourced MLLMs \cite{LLaVA-1.5, mPLUG-Owl2} based on the AesMMIT dataset, which not only enables the models to retain their original general knowledge but also facilitates the aesthetics perception capabilities, obtaining the multi-modality \textbf{Aes}thetic \textbf{Expert} models, dubbed \textbf{AesExpert}.

The contributions of this study are summarized as follows:


$\bullet$ \textbf{Aesthetic instruction-following dataset.} We construct a corpus-rich aesthetic critique database with 21,904 diverse-sourced images and 88K human natural language feedbacks to align MLLMs with human aesthetics perception. Further, we prompt GPT to refine the human aesthetic critiques and assemble the large-scale aesthetic instruction tuning dataset (AesMMIT) to ensure that MLLMs can handle diverse queries, which consists of 409K instructions covering multiple aesthetic perception dimensions, to activate stronger aesthetic capabilities of MLLMs.

$\bullet$ \textbf{AesExpert model.} We propose multi-modality aesthetic expert models with the aid of the proposed AesMMIT dataset via instruction fine-tuning. Extensive experiments demonstrate that the proposed AesExpert models deliver significantly better aesthetic perception performances than the state-of-the-art MLLMs, including the most advanced GPT-4V and Gemini-Pro-Vision.

$\bullet$ \textbf{Open source.} We release the following sources to the community: (1) the constructed aesthetic multi-modality instruction tuning dataset; (2) the proposed AesExpert model, including codes and checkpoints; (3) the visual AesExpert-Chatbot demo. We believe this work would shed light on building more advanced MLLMs with comprehensive aesthetic capabilities.

  \begin{figure*}[!t]
    \centerline{\includegraphics[width=\textwidth]{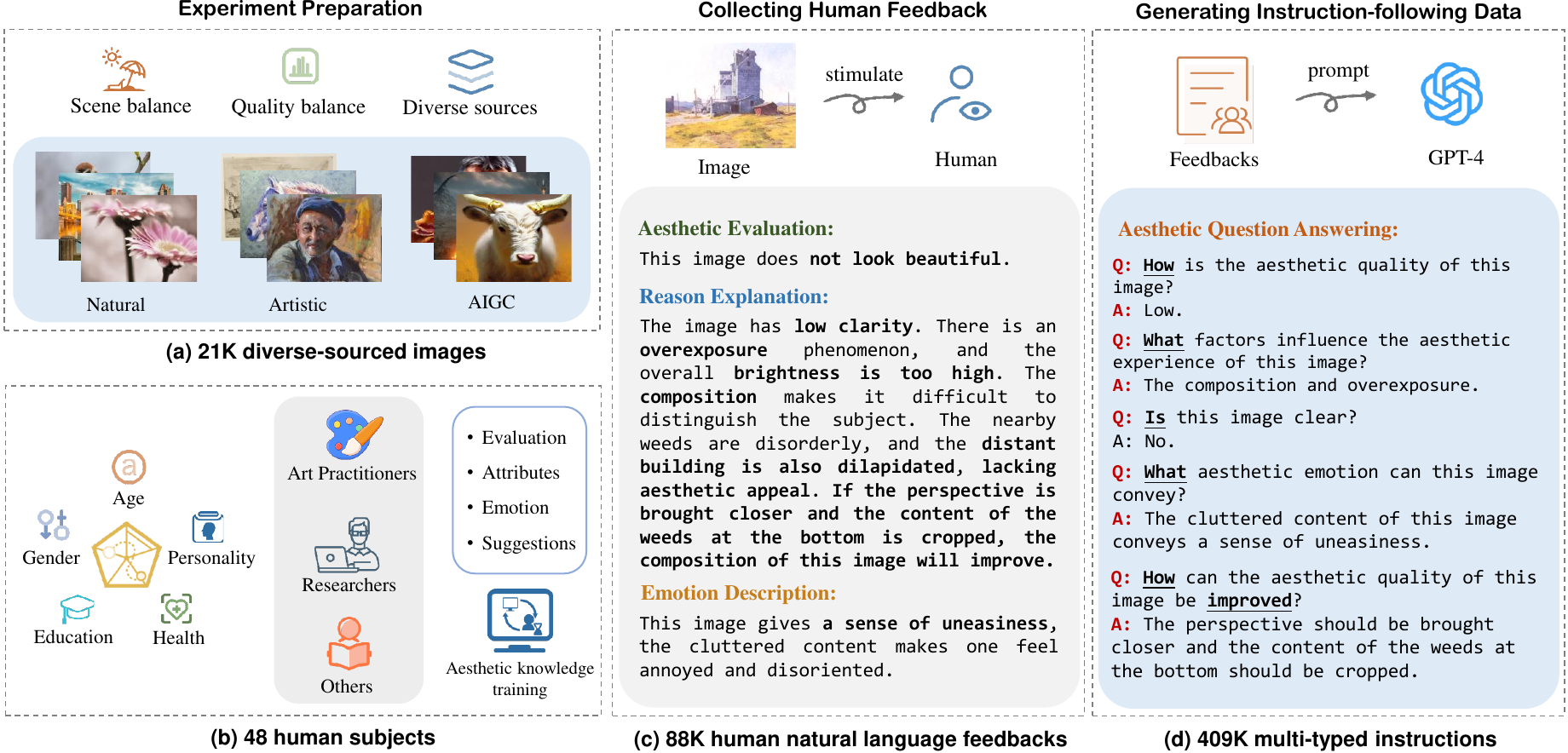}}
    \caption{The proposed dataset construction pipeline. First, we filter 21K diverse-sourced images based on scene, quality and source, and invite 48 human subjects who met the comprehensive criteria. Then, we collect 88K human feedbacks on image aesthetics perception. Finally, human feedbacks are converted into 409K instruction-following pairs (the AesMMIT dataset).
}
    \label{expert}
    \end{figure*}

\section{Related Work}

\subsection{Multi-modality Foundation Model}

Recently, large language models (LLMs), such as GPT-3 \cite{gpt4}, Flan-T5 \cite{Scaling} and LLaMA \cite{LLaMA}, have made remarkable progress in pure-textual tasks. The success of LLMs has also driven the research on vision-language interaction, resulting in the development of various multi-modality foundation models, \textit{e.g.} LLaVA \cite{LLAVA}, MiniGPT-4 \cite{MiniGPT-4}, mPLUG-Owl \cite{mPLUG-Owl}, Otter \cite{Otter} and Instruct-BLIP \cite{InstructBLIP}. These models typically contain a pre-trained visual encoder for image processing, an LLM for interpreting instructions and generating responses, and a cross-modality module to align the vision encoder with the LLM \cite{mi2024ze}. Despite their impressive performance in general-purpose visual tasks, their performance in handling image aesthetics perception remains underexplored. Therefore, \textbf{this work is dedicated to building the multi-modality aesthetic large model.}

\subsection{Multimodal Instruction-following Dataset}

Instruction tuning is an optimization method for multi-modality foundation models, aiming to improve their ability to perform specific tasks \cite{LLAVA}. With the development of MLLMs, researchers often employ ChatGPT \cite{gpt4} or GPT-4 \cite{GPT-4V} to generate diverse and expansive instruction-following data. For example, Zhu \textit{et al.} \cite{MiniGPT-4} employed GPT-3.5 to generate and improve detailed captions for high-quality instruction-following data (MiniGPT-4), which are mainly designed for general visual tasks. Liu \textit{et al.} \cite{LLAVA, LLaVA-1.5} proposed the multimodal instruction-following dataset (LLaVA-Instruct-150K) based on the existing COCO \cite{COCO} bounding box and caption dataset using GPT-4. In InstructBLIP \cite{InstructBLIP}, the authors transformed 13 vision-language tasks (\textit{e.g.} OCR-VQA \cite{OCR-VQA}) into the instruction-following format for instruction tuning. Chen \textit{et al.} \cite{ShareGPT4V} proposed a large-scale image-text dataset featuring 100K highly descriptive captions generated by GPT-4V and 1.2M high-quality captions generated by the proposed caption model, named ShareGPT4V. A more comprehensive survey can be found in \cite{Survey}.

The existing instruction tuning datasets are mainly constructed for general visual tasks, and the lack of image aesthetic data limits the aesthetic perception ability of MLLMs to a large extent. Further, the existing studies suggested that hallucination has become one of the most urgent problems for current MLLMs \cite{Hallucination1, Hallucination2}, therefore, using MLLM to directly generate instruction fine-tuning data for aesthetic tasks may further exacerbate this situation. To bridge this gap, this paper presents a corpus-rich aesthetic critique database by collecting human natural language feedback, based on which we establish a comprehensively annotated aesthetic multi-modality instruction tuning dataset, \textit{i.e.} AesMMIT. Further, we propose multi-modality aesthetic expert models based on aesthetic instruction fine-tuning, achieving significantly better aesthetic perception performances.

\section{Dataset Construction}
\label{Dataset}
In this section, we provide a detailed expatiation of the process in building the AesMMIT dataset, which is illustrated in Figure \ref{expert}. Specifically, subsection \ref{Experiment} describes the preparation of the subjective experiment including the collection of images and the recruitment of participants. Subsection \ref{Human} elaborates how we conduct the subjective experiment to obtain the 88K human feedbacks on 21,904 multi-sourced images. Subsection \ref{instruction} explains how we prompt GPT-4 to refine the aesthetic critiques and obtain the 409K instruction-following data covering multiple aesthetic perception dimensions and commonly-used question types. 

\subsection{Experiment Preparation}
\label{Experiment}

\textbf{Image Collection:} To guarantee the diversity of image types, we first collect a large number of images from various sources \cite{xu-4}, including \emph{natural images} (NIs), \emph{artistic images} (AIs) and \emph{artificial intelligence-generated images} (AGIs), as shown in Figure \ref{expert}(a). Then, we use a well-trained scene classification model \cite{CLIP} to automatically predict the scene label for each image, based on which we sampled these images to maintain scene diversity. To reduce the long-tail distribution of randomly sampled images \cite{Zhichao, xu-6}, we add 1239 high-aesthetic images from LITE \cite{LITE} and Impressions \cite{Impressions} datasets, and 1944 low-quality images from SPAQ \cite{SPAQ} and KonIQ-10K \cite{KonIQ-10K} datasets. Finally, a total of 21,904 images are collected, which are further sent to human subjects to collect aesthetic feedbacks based on subjective experiments. Detailed image source of the AesMMIT dataset is summarized in Table \ref{datasets}.

\textbf{Subject Selection:} To ensure the completeness of annotation and the diversity of corpus, we recruit subjects by considering five different perspectives, including \textit{age}, \textit{gender}, \textit{education}, \textit{health status} and \textit{personality}, as shown in Figure \ref{expert}(b). Specifically, we ensure that each subject is in good health and passes the Ishihara color blindness test \cite{PARA}. To enhance the quality of annotation, all subjects need to have a high school degree or above, and some experience in photography. Following the existing works \cite{xu-3, Q-Bench, xu-5}, to ensure the diversity of annotation, we further consider age distribution, gender balance, and personality diversity \cite{Zhu-TIP}. Finally, considering the difficulty of labeling artistic and artificial intelligence-generated images, we specially invited 6 art practitioners and 9 researchers in the field of image aesthetics evaluation to join the subjective experiment and focus on labeling these two kinds of images. In addition, we also organize two aesthetic knowledge training sessions before the experiment covering \textit{evaluation}, \textit{attribute}, \textit{emotion} and \textit{suggestion}, aiming to provide them with a more comprehensive understanding of image aesthetics annotation. Finally, based on the principle of voluntary participation \cite{PARA}, we invite 48 eligible human subjects to ensure the validity and reliability of annotations.

\subsection{Collecting Human Aesthetic Feedback}
\label{Human}

To enhance the understanding and interpretation capabilities of MLLMs in terms of aesthetics perception, we invite 48 subjects to contribute their insights and descriptions on image aesthetics through progressive questions, as illustrated in Figure \ref{expert}(c). This process includes three parts for each test image from coarse to fine. 


1) \textbf{Coarse-grained aesthetic evaluation.} The focus of this question is to collect simple aesthetic grade judgments on images. Participants could express their overall impression of image aesthetics using simple sentences such as \textit{good-looking}, \textit{beautiful}, \textit{average} or \textit{bad-looking}, \textit{etc}. Referring to common settings in the field of image quality assessment \cite{AesSurvey, AADB}, we recommend that experimental participants establish a uniform measure of five aesthetic grades and guarantee language diversity.

\begin{table}[!t]
\caption{Overview of the image source datasets.}  \label{datasets}
\renewcommand{\arraystretch}{1.5}
\fontsize{8.5pt}{7pt}\selectfont %
\centering
\setlength{\tabcolsep}{1.9mm}{
\begin{tabular}{c|lc}
\toprule
Type            &Dataset &Sampled Size
\\
      \midrule 
 \multirow{7}*{NIs}               
 &AADB \cite{AADB} &1096 \\
 &PARA \cite{PARA} &3694  \\
 &TAD66K \cite{TAD66K} &4003    \\
 &LITE \cite{LITE}&1087  \\
 &Impressions \cite{Impressions} &152    \\
 &SPAQ \cite{SPAQ} &1153   \\
 &KonIQ-10K \cite{KonIQ-10K} &791   \\
\midrule 
  \multirow{3}*{AIs}             
  &BAID \cite{BAID}&2970 \\
  &CAD \cite{CAD}&59   \\
  &ArtEmis \cite{ArtEmis}&1957  \\
  \midrule 
 \multirow{3}*{AGIs}
  &DiffusionDB \cite{DiffusionDB}  &4228 \\
 &AGIQA-3K \cite{AGIQA-1K}  &570   \\ 
 &AGIQA-1K \cite{AGIQA-3K} &144  \\
\bottomrule
\end{tabular}}
\end{table}

2) \textbf{Fine-grained reasoning and explanation.} On the basis of simple aesthetic grade judgments, this question is designed to mine the detailed reasons that influence human aesthetic judgments \cite{CADAS}. In this experiment, all participants need to explain what specific aspects of the image enhance or reduce the aesthetic appeal, such as \textit{clarity, color, light, image object} and \textit{composition}. In addition, we encourage participants to provide suggestions for the improvement of low-aesthetics images.

3) \textbf{Finer-grained description on aesthetic feeling.} This question breaks through the description of the inherent aesthetic attributes and aims to explore the impact of human emotions on aesthetic perception when viewing images. Participants need to describe some of the more abstract emotional factors in the image, \textit{e.g.} \textit{novel shooting view, interesting content, expressed emotions, etc}. Furthermore, we encourage participants to analyze the impact of these factors on image aesthetics.


The setting of the subjective experiment refers to the ITU-R Recommendation BT.500-13 standard \cite{ITU}. By systematically collecting feedback from the above three parts, we construct the corpus-rich aesthetic critique database, called \textbf{AesFeedback}, which contains \textbf{88K human feedbacks} on \textbf{21,904 multi-sourced images}. Since these annotations are all provided by human subjects, AesFeedback can capture the characteristics of basic aesthetic perceptions and the evaluation reasoning process, providing a valuable resource for improving the aesthetic capabilities of MLLMs.

  \begin{figure}[!t]
    \centerline{\includegraphics[width=7.5cm]{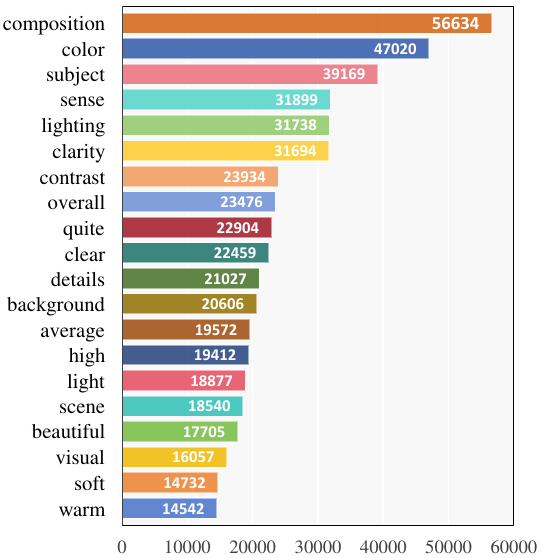}}
    \caption{Statistics of top-20 frequently-used words in the AesMMIT database.
}
    \label{words}
    \end{figure}
    
\subsection{Generating Instruction-following Data}
\label{instruction}
While the corpus-rich aesthetic critiques in the \textbf{AesFeedback} dataset can provide rich knowledge for aesthetic instruction tuning of MLLMs, we further design more instruction-following data to allow MLLMs to respond to a variety of human queries, achieving stronger aesthetic capabilities. Similar to existing works \cite{ShareGPT4V, Q-Instruct}, we leverage GPT-4 to transform human feedback into instruction-following formats, as illustrated in Figure \ref{expert}(d). Through this process, the proposed AesMMIT dataset includes 409K instruction-response pairs, with its details as follows.

\textbf{Aesthetic Description:} Similar to aesthetic caption \cite{AesCLIP}, the ability of image aesthetic description is crucial for MLLMs. As shown in Figure \ref{expert}(c), the collected AesFeedback dataset contains direct and comprehensive human natural language critiques on image aesthetics. Furthermore, these critiques provide aesthetic interpretation from overall aesthetic grades to fine-grained aesthetic attributes (\textit{e.g. clarity, color, light}, and \textit{image object}), which can activate the preliminary aesthetic interpretation abilities of MLLMs (see Figure \ref{comparison}). Therefore, we directly use the questions in the AesFeedback dataset as instructions and human critiques as responses, based on which we obtain the 88K instruction-following pairs of the proposed AesMMIT dataset.

\textbf{Aesthetic Question Answering:} In addition to directly adopting the AesFeedback as aesthetic instruction-following data, we further propose a GPT-assist approach to refine the aesthetic critiques and assemble a larger-scale aesthetic question-answering subset (AesVQA). Inspired by Q-Instruct \cite{Q-Instruct} and ShareGPT \cite{ShareGPT4V}, we prompt GPT-4 to generate diverse-style open-ended questions and provide corresponding brief answers, where both the questions and answers are based on human aesthetic descriptions. To ensure that MLLMs can handle diverse queries during the interaction, the questions are generated to cover commonly used question types. Specifically, \textbf{Yes-or-No-style} questions are straightforward queries that demand a simple \textit{yes} or \textit{no} as an answer, which are mainly used to provide a clear, binary response. \textbf{What-style} questions are leveraged to measure more comprehensive and complex aesthetic perception (\textit{e.g.} various aesthetic attributes). \textbf{How-style} questions are used to ask more details about aesthetic emotion, aesthetic attributes, and improvement suggestions; \textbf{Why-style} questions are employed to explore the foundational aspects of what makes images aesthetically appealing or unappealing, revealing the rationale of beauty and visual appeal. To ensure the diversity of instructions, following the existing works \cite{LLaVA-1.5, ScienceQA}, besides the direct answers, we also generate several wrong alternatives for the questions and convert them into an additional multi-choice question format \cite{co-instruct}. Through the above process, we obtain the final \textbf{AesMMIT} dataset, which consists of \textbf{409K multi-typed instructions}, aiming to activate stronger aesthetic capabilities. Figure \ref{words} shows the frequently occurring words in the proposed AesMMIT dataset. It can be observed that most nouns, adjectives and adverbs are related to image aesthetic description, which is quite different from common semantic-based tasks.


\section{Model Fine-tuning}


\definecolor{Ocean}{RGB}{129,194,234}

\subsection{Model Architecture}

The proposed AesExpert model follows the design of LLaVA-1.5 \cite{LLaVA-1.5}, which includes three components: (1) A visual model based on the CLIP-ViT-L14 \cite{CLIP} with an input size of 336$\times$336, which converts the input image into 576 tokens. (2) A visual-language projector based on two-layer multi-layer perception (MLP), which is employed to connect the visual modality and language modality. (3) A language model based on the open-sourced Vicuna-v1.5 \cite{Judging}, which is used to interpret instructions and generate responses. In this work, we build the AesExpert models based on two variants of LLaVA-1.5, including LLaVA-1.5-7B and LLaVA-1.5-13B. In addition, we also introduce the mPLUG-Owl2 \cite{mPLUG-Owl2} as the backbone to validate the aesthetic perception performance based on the AesMMIT dataset. The structures for these multi-modality foundation models are summarized in Table \ref{baseline}.

\begin{table}[!t]
\caption{Structures of the multi-modality foundation models for aesthetic instruction tuning.}  \label{baseline}
\renewcommand{\arraystretch}{1.5}
\fontsize{8.5pt}{7pt}\selectfont %
\centering
\setlength{\tabcolsep}{1.5mm}{
\begin{tabular}{c|lcc}
\toprule
Model &Visual Model &V$\rightarrow$L &Language Model \\
      \midrule 
            
LLaVA-1.5-7B &CLIP-ViT-L14 &MLP &Vicuna-v1.5-7B\\

LLaVA-1.5-13B  &CLIP-ViT-L14 &MLP &Vicuna-v1.5-13B \\

mPLUG-Owl2 &CLIP-ViT-L14 &Abstractor &LLaMA2-7B \\

\bottomrule
\end{tabular}}
\end{table}

\begin{table*}[!t]
\caption{Performance comparisons of the proposed AesExpert with existing MLLMs. AesA: Aesthetic Assessment, AesI: Aesthetic Interpretation, T. Q.: Technical quality, C. L.: Color and light, Comp.: Composition, Cont.: Content, Emot.: Emotion, Inte.: Interest, Uniq.: Uniqueness, L-13B: LLaVA-1.5-13B, Owl2: mPLUG-Owl2, L-7B: LLaVA-1.5-7B. The overall score represents the average of the four dimension scores.}  \label{mllms}
\renewcommand{\arraystretch}{1.5}
\fontsize{8.5pt}{7pt}\selectfont %
\centering
\setlength{\tabcolsep}{1mm}{
\begin{tabular}{l|cccc|c|cccc|c|c|c|cc}
\toprule
\multirow{2}*{Model}    & \multicolumn{5}{c|}{\textbf{Aesthetic Perception}}& \multicolumn{5}{c|}{\textbf{Aesthetic Empathy}} 
& \multicolumn{1}{c|}{\textbf{AesA}}  & \multicolumn{1}{c|}{\textbf{AesI}}  & {\multirow{2}*{\textbf{Overall}}}& {\multirow{2}*{\textbf{Rank}}}\\
\cmidrule(lr){2-6}\cmidrule(lr){7-11}\cmidrule(lr){12-12}\cmidrule(lr){13-13}
&\emph{T. Q.} &\emph{C. L.} &\emph{Comp.} &\emph{Cont.} &\emph{\textbf{Score}} &\emph{Emot.} &\emph{Inte.} &\emph{Uniq.} &\emph{Vibe} &\emph{\textbf{Score}} &\emph{\textbf{Score}}  &\emph{\textbf{Score}} & \\
\midrule 
\rowcolor{Ocean!60}
\textbf{AesExpert (L-13B)}      &{70.83\%} &\textbf{80.63\%} &\textbf{81.75\%} &\textbf{76.60\%}   &\textbf{79.54\%}    &\textbf{82.94\%} &80.65\% &\textbf{93.10\%} &\textbf{88.56\%}    &\textbf{84.89\%}    &\textbf{59.57\%}    &\textbf{1.340}    &\textbf{89.50\%}    &\textbf{1} \\
\rowcolor{Ocean!40}
{AesExpert (Owl2)}      &{70.19\%} &{79.47\%} &{79.56\%} &{71.28\%}   &{77.64\%}    &{80.66\%} &83.87\% &{89.66\%} &{86.31\%}    &{82.68\%}    &{56.32\%}    &{1.336}    &{87.56\%}    &{2} \\
\rowcolor{Ocean!20}
{AesExpert (L-7B)}      &\textbf{72.44\%} &{79.38\%} &{80.66\%} &{74.47\%}   &{78.57\%}    &{82.05\%} &83.87\% &{86.21\%} &{87.81\%}    &{84.04\%}    &{53.25\%}    &{1.317}    &{86.89\%}    &{3} \\
GPT-4V                  &69.02\% &74.66\% &71.72\% &65.57\%   &72.08\%    &65.06\% &72.41\% &62.07\% &80.15\%    &70.16\%    &50.86\%    &1.301    &80.80\%    &4   \\
ShareGPT4V              &62.18\% &71.90\% &69.29\% &64.89\%   &69.18\%    &66.48\% &80.65\% &68.97\% &78.72\%    &70.75\%    &47.82\%    &1.296    &79.34\%    &5   \\
LLaVA-1.5-13B           &67.63\% &74.65\% &70.09\% &68.44\%   &71.61\%    &67.15\% &80.65\% &75.86\% &81.18\%    &72.07\%    &49.82\%    &1.222    &78.93\%    &6   \\
Gemini Pro Vision       &65.08\% &74.57\% &72.24\% &67.97\%   &71.99\%    &66.87\% &\textbf{87.50\%} &70.00\%    &79.09\%    &71.37\%    &49.38\%  &1.222    &78.74\%    &7  \\
mPLUG-Owl2              &60.90\% &70.57\% &68.30\% &62.77\%   &67.89\%    &65.60\% &77.42\% &65.52\% &78.07\%    &69.89\%    &50.57\%    &1.182    &76.64\%    &8   \\
Q-Instruct              &66.03\% &74.48\% &73.68\% &68.09\%   &72.61\%    &68.64\% &83.86\% &75.86\% &80.00\%    &72.68\%    &52.86\%    &1.020    &75.04\%    &9   \\
LLaVA-1.5-7B            &53.85\% &70.16\% &67.40\% &59.93\%   &66.32\%    &62.49\% &80.65\% &75.85\% &78.93\%    &68.32\%    &45.46\%    &1.157    &73.95\%    &10   \\
Qwen-VL                 &54.81\% &66.25\% &62.91\% &60.64\%   &63.21\%    &58.67\% &83.87\% &72.41\% &73.90\%    &64.18\%    &46.25\%    &1.192    &73.21\%    &11  \\
LLaVA                   &46.79\% &63.59\% &65.30\% &64.54\%   &62.43\%    &58.61\% &80.63\% &65.52\% &75.83\%    &64.68\%    &45.96\%    &1.125    &71.39\%    &12   \\
InstructBLIP            &37.82\% &55.36\% &55.43\% &57.09\%   &54.29\%    &49.64\% &58.06\% &51.72\% &61.50\%    &53.89\%    &46.54\%    &1.126    &66.83\%    &13   \\
IDEFICS-Instruct        &37.50\% &52.87\% &52.84\% &51.06\%   &50.82\%    &43.93\% &64.52\% &62.07\% &64.06\%    &50.82\%    &45.00\%    &1.180    &66.16\%    &14   \\
Otter                   &35.90\% &54.28\% &51.65\% &51.06\%   &50.96\%    &48.42\% &70.97\% &51.72\% &63.21\%    &53.64\%    &44.86\%    &1.027    &63.04\%    &15 \\
MiniGPT-v2              &56.73\% &56.44\% &51.74\% &50.00\%   &54.18\%    &52.52\% &58.06\% &44.83\% &58.07\%    &54.36\%    &31.11\%    &1.003    &59.99\%    &16   \\
GLM                     &55.77\% &54.61\% &51.25\% &48.94\%   &52.96\%    &53.13\% &70.97\% &44.83\% &55.29\%    &53.96\%    &37.79\%    &0.932    &59.48\%    &17   \\
MiniGPT-4               &39.42\% &41.31\% &42.67\% &44.33\%   &41.93\%    &39.78\% &38.71\% &24.14\% &39.04\%    &39.35\%    &38.57\%    &0.999    &54.94\%    &18   \\
TinyGPT-V               &21.79\% &24.52\% &22.13\% &28.01\%   &23.71\%    &30.36\% &29.03\% &31.03\% &35.40\%    &32.04\%    &43.57\%    &0.701    &42.36\%    &19   \\

\bottomrule
\end{tabular}}
\end{table*}

\subsection{Supervised Fine-Tuning}

In general, the training of open-sourced MLLMs \cite{LLAVA, LLaVA-1.5, MiniGPT-4} includes two stages: aligning the representation space of the visual backbone and the LLM with million-scale web data \cite{ScienceQA}, and visual instruction tuning with a combination of multi-modality datasets \cite{ShareGPT4V, InstructBLIP}. Considering that our purpose is to improve the aesthetic perception ability of the current MLLMs, we directly use the proposed AesMMIT dataset to perform supervised instruction fine-tuning on the models pre-trained on general-purpose visual tasks \cite{LLaVA-1.5}. Following existing works \cite{Q-Instruct, MiniGPT-v2}, supervised instruction fine-tuning not only enables the models to retain their original general knowledge but also facilitates the aesthetics perception capabilities. In this work, to enhance the training efficiency and compare fairly, we freeze the vision model and focus on fine-tuning the projector and the language model. Based on supervised instruction fine-tuning, we implemented three different versions of multi-modality aesthetic expert models.

\section{Experiments}

 \subsection{Implementation Details}
\textbf{Training setting}: In this work, we finetune three pre-trained MLLMs based on the constructed AesMMIT dataset in full schedule mode, including LLaVA-1.5-7B (Vicuna-v1.5-7B) \cite{LLaVA-1.5}, LLaVA-1.5-13B (Vicuna-v1.5-13B) \cite{LLaVA-1.5} and mPLUG-Owl2 (LLaMA-2-7B) \cite{mPLUG-Owl2}. In implementation, to ensure fairness, we follow the default hyper-parameters provided by the original models. All models are trained based on the GPU server with 8 NVIDIA Tesla A100 80G, and the evaluation experiments are conducted on the GPU server with 2 NVIDIA RTX 4090 24G.

\textbf{Benchmark}: To verify the performance of our proposed AesExpert models on image aesthetics perception, extensive experiments and comparisons are conducted on the AesBench \cite{AesBench}, which is a well-designed benchmark for MLLMs on aesthetics perception evaluation. Specifically, AesBench contains 2,800 images and four evaluation criteria designed from four dimensions, including (1) \textbf{Aesthetic Perception} (\emph{AesP}) focuses on the ability of MLLMs to recognize and understand aesthetic attributes. (2) \textbf{Aesthetic Empathy} (\emph{AesE}) evaluates the ability of MLLMs to resonate with the emotional aspects conveyed through aesthetic expressions like humans. (3) \textbf{Aesthetic Assessment} (\emph{AesA}) evaluates the ability of MLLMs to judge image aesthetics quality. (4) \textbf{Aesthetic Interpretation} (\emph{AesI}) involves the ability of MLLMs to interpret and analyze the reasons for aesthetic quality. Each dimension contains 2,800 questions and correct answers. For the first three dimensions, the accuracy of the answers is used to measure the performance of the model, while the AesI is evaluated based on GPT scoring.

\begin{table*}[!t]
\caption{Comparison of the \textbf{Aesthetic \underline{Perception}} ability between baseline MLLMs and the proposed AesExpert models.}  \label{EAesP}
\renewcommand{\arraystretch}{1.5}
\fontsize{8.5pt}{7pt}\selectfont %
\centering
\setlength{\tabcolsep}{1mm}{
\begin{tabular}{l|cccc|ccc|cccc|cc}
\toprule
\multirow{2}*{MLLM}    & \multicolumn{4}{c|}{\textbf{Perceptual Dimensions}}& \multicolumn{3}{c|}{\textbf{Image Sources}} 
& \multicolumn{4}{c|}{\textbf{Question Types}}   & {\multirow{2}*{\textbf{Overall}}}\\
\cmidrule(lr){2-5}\cmidrule(lr){6-8}\cmidrule(lr){9-12}
&\emph{Tec. Qua.} &\emph{Col. Lig.} &\emph{Composition} &\emph{Content} &\emph{NIs} &\emph{AIs} &\emph{AGIs }  &\emph{Yes-No } &\emph{What} &\emph{How} &\emph{Why}  \\
\midrule 
BL (mPLUG-Owl2)              &60.90\% &70.57\% &68.30\% &62.77\%   &72.23\% &64.71\% &64.10\%    &65.59\% &58.64\% &73.02\% &80.73\%       &67.89\% \\
\textbf{AesExpert}              &\textbf{70.19\%} &\textbf{79.47\%}&\textbf{79.56\%}&\textbf{71.28\%}&\textbf{79.78\%}&\textbf{73.32\%}&\textbf{78.72\%}&\textbf{72.75\%}&\textbf{66.43\%}&\textbf{89.32\%}&\textbf{89.02\%}&\textbf{77.64\%} \\
\rowcolor{lightgray!25}
Improvement              &\textcolor[RGB]{0, 27, 197}{+9.29\%} &\textcolor[RGB]{0, 27, 197}{+8.90\%}&\textcolor[RGB]{0, 27, 197}{+11.26\%}&\textcolor[RGB]{0, 27, 197}{+8.51\%}&\textcolor[RGB]{0, 27, 197}{+7.55\%}&\textcolor[RGB]{0, 27, 197}{+8.61\%}&\textcolor[RGB]{0, 27, 197}{+14.62\%}&\textcolor[RGB]{0, 27, 197}{+7.16\%}&\textcolor[RGB]{0, 27, 197}{+7.79\%}&\textcolor[RGB]{0, 27, 197}{+16.30\%}&\textcolor[RGB]{0, 27, 197}{+8.29\%}&\textcolor[RGB]{0, 27, 197}{+9.75\%}\\

BL (LLaVA-1.5-7b)               &53.85\% &70.16\% &67.40\% &59.93\%   &69.10\% &65.71\% &62.37\%    &62.36\% &58.92\% &70.71\% &81.22\%       &66.32\%\\
\textbf{AesExpert}              &\textbf{72.44\%} &\textbf{79.38\%} &\textbf{80.66\%} &\textbf{74.47\%}   &\textbf{81.46\%} &\textbf{74.69\%} &\textbf{77.93\%}    &\textbf{70.74\%} &\textbf{72.52\%} &\textbf{90.33\%} &\textbf{88.05\%}      &\textbf{78.57\%} \\
\rowcolor{lightgray!25}
Improvement              &\textcolor[RGB]{0, 27, 197}{+18.59\%} &\textcolor[RGB]{0, 27, 197}{+9.22\%}&\textcolor[RGB]{0, 27, 197}{+13.26\%}&\textcolor[RGB]{0, 27, 197}{+14.54\%}&\textcolor[RGB]{0, 27, 197}{+12.36\%}&\textcolor[RGB]{0, 27, 197}{+8.98\%}&\textcolor[RGB]{0, 27, 197}{+15.56\%}&\textcolor[RGB]{0, 27, 197}{+8.38\%}&\textcolor[RGB]{0, 27, 197}{+13.60\%}&\textcolor[RGB]{0, 27, 197}{+19.62\%}&\textcolor[RGB]{0, 27, 197}{+6.83\%}&\textcolor[RGB]{0, 27, 197}{+12.25\%}\\

BL (LLaVA-1.5-13b)               &67.63\% &74.65\% &70.09\% &68.44\%   &75.36\% &70.32\% &66.76\%    &68.72\% &61.19\% &78.21\% &85.37\%       &71.61\%\\
\textbf{AesExpert}              &\textbf{70.83\%} &\textbf{80.63\%} &\textbf{81.75\%} &\textbf{76.60\%}   &\textbf{81.54\%} &\textbf{76.93\%} &\textbf{78.99\%}    &\textbf{71.14\%} &\textbf{70.40\%} &\textbf{93.80\%} &\textbf{91.46\%}      &\textbf{79.54\%} \\

\rowcolor{lightgray!25}
Improvement              &\textcolor[RGB]{0, 27, 197}{+3.20\%} &\textcolor[RGB]{0, 27, 197}{+5.98\%}&\textcolor[RGB]{0, 27, 197}{+11.66\%}&\textcolor[RGB]{0, 27, 197}{+8.16\%}&\textcolor[RGB]{0, 27, 197}{+6.18\%}&\textcolor[RGB]{0, 27, 197}{+6.61\%}&\textcolor[RGB]{0, 27, 197}{+12.23\%}&\textcolor[RGB]{0, 27, 197}{+2.42\%}&\textcolor[RGB]{0, 27, 197}{+9.21\%}&\textcolor[RGB]{0, 27, 197}{+15.59\%}&\textcolor[RGB]{0, 27, 197}{+6.09\%}&\textcolor[RGB]{0, 27, 197}{+7.93\%}\\

\bottomrule
\end{tabular}}
\end{table*}

\begin{table*}[!t]
\caption{Comparison of the \textbf{Aesthetic \underline{Empathy}} ability between baseline MLLMs and the proposed AesExpert models.}  \label{EAesE}
\renewcommand{\arraystretch}{1.5}
\fontsize{8.5pt}{7pt}\selectfont %
\centering
\setlength{\tabcolsep}{1mm}{
\begin{tabular}{l|cccc|ccc|cccc|cc}
\toprule
\multirow{2}*{MLLM}    & \multicolumn{4}{c|}{\textbf{Empathy Dimensions}}& \multicolumn{3}{c|}{\textbf{Image Sources}} 
& \multicolumn{4}{c|}{\textbf{Question Types}}   & {\multirow{2}*{\textbf{Overall}}}\\
\cmidrule(lr){2-5}\cmidrule(lr){6-8}\cmidrule(lr){9-12}
&\emph{Emotion} &\emph{Interest} &\emph{Uniqueness} &\emph{Vibe}&\emph{NIs} &\emph{AIs} &\emph{AGIs }  &\emph{Yes-No } &\emph{What} &\emph{How} &\emph{Why}  \\
\midrule 
BL (mPLUG-Owl2)           &65.60\% &77.42\% &65.52\% &78.07\%    &71.03\% &71.57\% &66.22\%    &{68.05\%} &64.16\% &70.14\% &83.82\%       &69.89\% \\
\textbf{AesExpert}              &\textbf{80.66\%} &\textbf{83.87\%}&\textbf{89.66\%}&\textbf{86.31\%}&\textbf{84.27\%}&\textbf{80.42\%}&\textbf{82.45\%}&\textbf{72.52\%}&\textbf{80.59\%}&\textbf{94.29\%}&\textbf{90.93\%}&\textbf{82.68\%} \\
\rowcolor{lightgray!25}
Improvement              &\textcolor[RGB]{0, 27, 197}{+15.06\%} &\textcolor[RGB]{0, 27, 197}{+6.45\%}&\textcolor[RGB]{0, 27, 197}{+24.14\%}&\textcolor[RGB]{0, 27, 197}{+8.24\%}&\textcolor[RGB]{0, 27, 197}{+13.24\%}&\textcolor[RGB]{0, 27, 197}{+8.85\%}&\textcolor[RGB]{0, 27, 197}{+16.23\%}&\textcolor[RGB]{0, 27, 197}{+4.47\%}&\textcolor[RGB]{0, 27, 197}{+16.43\%}&\textcolor[RGB]{0, 27, 197}{+24.15\%}&\textcolor[RGB]{0, 27, 197}{+7.11\%}&\textcolor[RGB]{0, 27, 197}{+12.79\%}\\

BL (LLaVA-1.5-7b)                &62.49\% &80.65\% &{75.85\%} &78.93\%    &69.26\% &69.58\% &65.43\%    &62.37\% &64.16\% &71.71\% &84.07\%       &68.32\%\\
\textbf{AesExpert}            &\textbf{82.05\%} &\textbf{83.87\%} &\textbf{86.21\%} &\textbf{87.81\%}    &\textbf{86.36\%} &\textbf{82.54\%} &\textbf{81.78\%}    &\textbf{74.04\%} &\textbf{83.14\%} &\textbf{95.29\%} &\textbf{90.44\%}      &\textbf{84.04\%} \\
\rowcolor{lightgray!25}
Improvement              &\textcolor[RGB]{0, 27, 197}{+19.56\%} &\textcolor[RGB]{0, 27, 197}{+3.22\%}&\textcolor[RGB]{0, 27, 197}{+10.36\%}&\textcolor[RGB]{0, 27, 197}{+8.88\%}&\textcolor[RGB]{0, 27, 197}{+17.10\%}&\textcolor[RGB]{0, 27, 197}{+12.96\%}&\textcolor[RGB]{0, 27, 197}{+16.35\%}&\textcolor[RGB]{0, 27, 197}{+11.67\%}&\textcolor[RGB]{0, 27, 197}{+18.98\%}&\textcolor[RGB]{0, 27, 197}{+23.58\%}&\textcolor[RGB]{0, 27, 197}{+6.37\%}&\textcolor[RGB]{0, 27, 197}{+15.72\%}\\

BL (LLaVA-1.5-13b)                &67.15\% &80.65\% &75.86\% &81.18\%    &72.79\% &74.44\% &68.35\%    &70.28\% &65.16\% &73.86\% &85.29\%       &72.07\%\\
\textbf{AesExpert}            &\textbf{82.94\%} &\textbf{80.65\%} &\textbf{93.10\%} &\textbf{88.56\%}    &\textbf{86.20\%} &\textbf{84.04\%} &\textbf{83.64\%}    &\textbf{74.14\%} &\textbf{84.84\%} &\textbf{95.57\%} &\textbf{92.65\%}      &\textbf{84.89\%} \\

\rowcolor{lightgray!25}
Improvement              &\textcolor[RGB]{0, 27, 197}{+15.79\%} &\textcolor[RGB]{0, 27, 197}{+0\%}&\textcolor[RGB]{0, 27, 197}{+17.24\%}&\textcolor[RGB]{0, 27, 197}{+7.38\%}&\textcolor[RGB]{0, 27, 197}{+13.41\%}&\textcolor[RGB]{0, 27, 197}{+9.60\%}&\textcolor[RGB]{0, 27, 197}{+15.29\%}&\textcolor[RGB]{0, 27, 197}{+3.86\%}&\textcolor[RGB]{0, 27, 197}{+19.68\%}&\textcolor[RGB]{0, 27, 197}{+21.71\%}&\textcolor[RGB]{0, 27, 197}{+7.36\%}&\textcolor[RGB]{0, 27, 197}{+12.82\%}\\

\bottomrule
\end{tabular}}
\end{table*}

\subsection{Performance Comparison}
In this section, we compare the performance of our proposed AesExpert models (three versions) with 16 state-of-the-art MLLMs, including the popular GPT-4V \cite{GPT-4V} and Gemini Pro Vision \cite{Gemini}, as well as 14 state-of-the-art variants with open sources, \emph{i.e.} LLaVA (LLaMA-2-Chat-7B) \cite{LLAVA}, LLaVA-1.5-7B (Vicuna-v1.5-7B) \cite{LLaVA-1.5}, LLaVA-1.5-13B (Vicuna-v1.5-13B) \cite{LLaVA-1.5}, ShareGPT4V (Vicuna-v1.5-7B) \cite{ShareGPT4V}, Q-Instruct (LLaVA-v1.5-7B) \cite{Q-Instruct}, mPLUG-Owl2 (LLaMA-2-7B) \cite{mPLUG-Owl2}, InstructBLIP (Vicuna-7B) \cite{InstructBLIP}, IDEFICS-Instruct (LLaMA-7B) \cite{IDEFICS}, Otter (MPT-7B) \cite{Otter}, Qwen-VL (QWen-7B) \cite{Qwen-VL}, GLM (ChatGLM-6B) \cite{GLM}, MiniGPT-v2 (LLaMA-2-Chat-7B) \cite{MiniGPT-v2}, TinyGPT-V (Phi-2) \cite{TinyGPT-V} and MiniGPT-4 (Vicuna-7B) \cite{MiniGPT-4}. Detailed information about these models can be found in \cite{Survey}. The experimental results are listed in Table \ref{mllms}.

From Table \ref{mllms}, we can find that the three versions of AesExpert achieve the top three results. Among them, AesExpert based on LLaVA-1.5-13B achieves the best performance, which is significantly better than the most advanced GPT-4V. For the existing open-sourced models, ShareGPT4V \cite{ShareGPT4V} performs best, but lags behind our AesExpert (L-13B) by more than 10\% on overall score. These experimental results reveal that the proposed AesExpert models have the best aesthetic perception abilities and highlight the advantage of the constructed AesMMIT dataset for improving multi-modality foundation models.

\subsection{Performance Improvement}
In this section, we quantitatively evaluate the aesthetic perception abilities of MLLMs after aesthetic instruction tuning in the four tasks defined by AesBench \cite{AesBench}. The experimental results are summarized in Tables \ref{EAesP}, \ref{EAesE}, \ref{EAesA} and \ref{EAesI}, respectively.

\begin{table}[!t]
\caption{Comparison of the \textbf{Aesthetic \underline{Assessment}} ability between baseline MLLMs and the proposed AesExpert models.} \label{EAesA}
\renewcommand{\arraystretch}{1.6}
\fontsize{8.5pt}{7pt}\selectfont %
\centering
\setlength{\tabcolsep}{1.3mm}{
\begin{tabular}{l|ccc|ccc|c}
\toprule
MLLM     &\emph{NIs} &\emph{AIs} &\emph{AGIs }  &\textbf{Overall}\\
      \midrule 
BL (mPLUG-Owl2)                &57.78\% &49.50\% &40.83\%    &50.57\%    \\
\textbf{AesExpert}                     &\textbf{64.45\%} &\textbf{51.50\%}&\textbf{48.01\%}&\textbf{56.32\%}    \\
\rowcolor{lightgray!25}
Improvement              &\textcolor[RGB]{0, 27, 197}{+6.67\%} &\textcolor[RGB]{0, 27, 197}{+2.00\%}&\textcolor[RGB]{0, 27, 197}{+7.18\%}&\textcolor[RGB]{0, 27, 197}{5.75\%}\\

BL (LLaVA-1.5-7b)                   &50.08\% &\textbf{48.13\%} &34.97\%    &45.46\%     \\
\textbf{AesExpert}             &\textbf{61.40\%} &{47.63\%} &\textbf{45.74\%}  &\textbf{53.25\%}     \\
\rowcolor{lightgray!25}
Improvement             &\textcolor[RGB]{0, 27, 197}{+11.32\%} &\textcolor[RGB]{0, 27, 197}{-0.50\%}&\textcolor[RGB]{0, 27, 197}{+10.77\%}&\textcolor[RGB]{0, 27, 197}{+7.79\%}\\

BL (LLaVA-1.5-13b)                   &56.66\% &49.63\% &38.70\%    &49.82\%     \\
\textbf{AesExpert}             &\textbf{67.82\%} &\textbf{53.74\%} &\textbf{52.13\%}  &\textbf{59.57\%}     \\
\rowcolor{lightgray!25}
Improvement             &\textcolor[RGB]{0, 27, 197}{+11.16\%} &\textcolor[RGB]{0, 27, 197}{+4.11\%}&\textcolor[RGB]{0, 27, 197}{+13.43\%}&\textcolor[RGB]{0, 27, 197}{+9.75\%}\\

\bottomrule
\end{tabular}}
\end{table}

\textbf{Aesthetic Perception Ability.} From Table \ref{EAesP}, we can observe that fine-tuning baseline MLLMs using AesMMIT can significantly improve their image aesthetic perception abilities. Specifically, among the three baseline MLLMs, LLaVA-1.5-7b achieves the most performance improvements (over 12\% on the overall score). For the four different perception dimensions, we noticed the most significant improvement for \textbf{composition}. The possible reason is that composition is a very important element in human aesthetic perception, which also can be found in Figure \ref{words}, indicating that composition is the most frequently occurring word in the AesMMIT dataset. Therefore, the proposed AesExpert model fine-tuned on AesMMIT has excellent composition perception capabilities. In addition, among the three types of images, \textbf{artificial intelligence-generated images} obtain the biggest performance improvement. This is mainly because that existing instruction fine-tuning datasets usually contain very few artificial intelligence-generated images, and our AesMMIT dataset makes up for this shortcoming, achieving significant performance improvement. Finally, the biggest performance improvement of the four question types is `\textit{How}'. These findings inspire us to further expand our dataset to cover more perception dimensions and more question types in future studies.

\textbf{Aesthetic Empathy Ability.} From Table \ref{EAesE}, it is observed that our AesExpert is superior to the baseline models by a large margin (more than 12\%), especially on LLaVA-1.5-7B, with performance improvement up to 15.72\%. Among the four dimensions, \textbf{Uniqueness} usually has significant improvement, indicating that the major concerns for empathy raised by humans in the AesMMIT dataset are related to the uniqueness of perspective. Among the three types of images, \textbf{artificial intelligence-generated images} still obtain the biggest performance improvement (more than 15\%). In addition, for the four question types, \textit{HOW} questions achieve the biggest performance improvement. In summary, these results underscore that AesMMIT can significantly improve the aesthetic empathy abilities of MLLMs, and even our AesExpert based on the LLaVA-1.5-7B model markedly surpasses the current top-performing GPT-4V (refer to Table \ref{mllms}).

\textbf{Aesthetic Assessment Ability.} The observations from Table \ref{EAesA} underscore that the aesthetic instruction tuning also notably improves the aesthetic assessment ability of MLLMs, especially on the \textit{artificial intelligence-generated images}. In contrast, the average improvement on \textit{artistic image} is less significant, implying that, due to the highly abstract nature, the aesthetic assessment of artistic images is still a relatively difficult task. We look forward to better solutions for artistic image in the future.

\begin{table}[t]
\caption{Comparison of the Aesthetic \underline{Interpretation} ability between baseline MLLMs and the proposed AesExpert models. Rele.: Relevance, Prec.: Precision, Comp.: Completeness. } \label{EAesI}
\renewcommand{\arraystretch}{1.6}
\fontsize{8.5pt}{7pt}\selectfont %
\centering
\setlength{\tabcolsep}{1.9mm}{
\begin{tabular}{l|ccc|c}
\toprule
MLLM  &\emph{Rele.} &\emph{Prec.} &\emph{Comp.} &\textbf{Overall}\\
\midrule 
BL (mPLUG-Owl2)                 &1.402 &1.016 &1.130 &1.182     \\
\textbf{AesExpert}                     &\textbf{1.406} &\textbf{1.431}&\textbf{1.171} &\textbf{1.336}    \\
\rowcolor{lightgray!25}
Improvement              &\textcolor[RGB]{0, 27, 197}{+0.40\%} &\textcolor[RGB]{0, 27, 197}{+41.50\%}&\textcolor[RGB]{0, 27, 197}{+4.10\%}&\textcolor[RGB]{0, 27, 197}{+15.40\%}\\

BL (LLaVA-1.5-7b)                     &\textbf{1.397} &0.953 &1.120 &1.157     \\
\textbf{AesExpert}                     &1.379 &\textbf{1.399} &\textbf{1.171} &\textbf{1.317}    \\
\rowcolor{lightgray!25}
Improvement              &\textcolor[RGB]{0, 27, 197}{-1.80\%} &\textcolor[RGB]{0, 27, 197}{+44.60\%}&\textcolor[RGB]{0, 27, 197}{+5.10\%}&\textcolor[RGB]{0, 27, 197}{+16.00\%}\\

BL (LLaVA-1.5-13B)                     &\textbf{1.403} &1.150 &1.113 &1.222     \\
\textbf{AesExpert}                     &1.412 &\textbf{1.409} &\textbf{1.198} &\textbf{1.340}    \\
\rowcolor{lightgray!25}
Improvement              &\textcolor[RGB]{0, 27, 197}{+0.90\%} &\textcolor[RGB]{0, 27, 197}{+25.90\%}&\textcolor[RGB]{0, 27, 197}{+8.50\%}&\textcolor[RGB]{0, 27, 197}{+11.80\%}\\

\bottomrule
\end{tabular}}
\end{table}

\textbf{Aesthetic Interpretation Ability}. Hallucination has been regarded as one of the critical challenges for MLLMs \cite{Hallucination1, Hallucination2}, which imagines incorrect details about an image in visual question answering. To alleviate this problem, the proposed AesMMIT dataset is collected from \textbf{human natural language feedback} rather than machine-generated annotations. As can be seen from Table \ref{EAesI}, the \textbf{Precision} of interpretation has been significantly improved. For three different baseline MLLMs, significant improvements of 41.50\%, 44.60\% and 25.90\% have been achieved improved, respectively. This result proves that the proposed AesMMIT allows the MLLMs to learn the style of human language and enhance the precision of aesthetic descriptions. In addition, from the overall score, the aesthetic instruction tuning based on AesMMIT significantly improves the aesthetic interpretation ability of MLLMs, especially for LLaVA-1.5-7B and mPLUG-Owl2 with performance improvement of 16.00\% and 15.40\%. These results demonstrate that the AesMMIT dataset could significantly benefit the existing MLLMs for obtaining enhanced aesthetic interpretation ability.

\subsection{Comparison of Training Data}

\begin{figure}[!t]
    \centerline{\includegraphics[width=8cm]{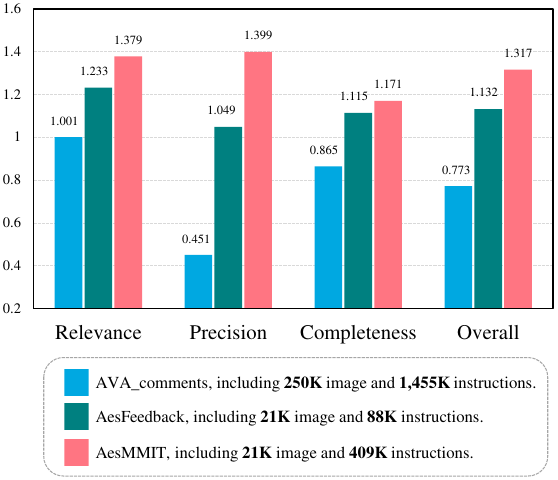}}
    \caption{Performance comparison using different datasets.}
    \label{comparison}
    \end{figure}

To verify the effectiveness and necessity of collecting human feedback for improving the aesthetic perception abilities of MLLMs, we further conduct experiments to compare our dataset with the AVA-Comments dataset \cite{AVA}, the largest multi-modality dataset in image aesthetics assessment, which contains over 250K images with 1,455K comments. Considering that the AVA-Comments dataset only contains aesthetic descriptions, for fairness, we evaluate the aesthetic interpretation ability across different datasets using the same instruction settings. In addition, we use the AesFeedback subset and the whole AesMMIT dataset to fine-tune the model for comparison, respectively. Figure \ref{comparison} provides the experimental results, where all experiments adopt the same LLaVA-1.5-7B as the baseline MLLM.

It can be observed from Figure \ref{comparison} that although AVA-Comments contains more images and instructions, it cannot provide effective aesthetic information to MLLM, resulting in poor aesthetic interpretation ability, especially on \textbf{Precision}. In contrast, the human feedback we collected (AesFeedback) can achieve pretty good aesthetic interpretation abilities for MLLMs. More importantly, the proposed AesMMIT dataset expanded by GPT can further improve the model performance. These results clearly demonstrate the effectiveness and necessity of the proposed AesMMIT dataset.

\section{Conclusion}

In this work, we have made an attempt to exploit the aesthetic perception ability of the multi-modality foundation model. Specifically, we first build a corpus-rich aesthetic critique database via human natural language feedback (\textbf{AesFeedback}), based on which we further establish a comprehensively annotated aesthetic multi-modality instruction tuning dataset (\textbf{AesMMIT}). In addition, we propose multi-modality aesthetic expert models based on aesthetic instruction fine-tuning, achieving significantly better aesthetic perception performances. We believe this work is a solid step in improving the aesthetic perception ability of MLLM, and we hope that our contribution will encourage the research community to build multi-modality foundation models that can understand highly abstract image aesthetics like humans.


\bibliographystyle{ACM-Reference-Format}
\bibliography{sample-base}


\begin{thebibliography}{73}


\ifx \showCODEN    \undefined \def \showCODEN     #1{\unskip}     \fi
\ifx \showDOI      \undefined \def \showDOI       #1{#1}\fi
\ifx \showISBNx    \undefined \def \showISBNx     #1{\unskip}     \fi
\ifx \showISBNxiii \undefined \def \showISBNxiii  #1{\unskip}     \fi
\ifx \showISSN     \undefined \def \showISSN      #1{\unskip}     \fi
\ifx \showLCCN     \undefined \def \showLCCN      #1{\unskip}     \fi
\ifx \shownote     \undefined \def \shownote      #1{#1}          \fi
\ifx \showarticletitle \undefined \def \showarticletitle #1{#1}   \fi
\ifx \showURL      \undefined \def \showURL       {\relax}        \fi
\providecommand\bibfield[2]{#2}
\providecommand\bibinfo[2]{#2}
\providecommand\natexlab[1]{#1}
\providecommand\showeprint[2][]{arXiv:#2}

\bibitem[Achlioptas et~al\mbox{.}(2021)]%
        {ArtEmis}
\bibfield{author}{\bibinfo{person}{Panos Achlioptas}, \bibinfo{person}{Maks Ovsjanikov}, \bibinfo{person}{Kilichbek Haydarov}, \bibinfo{person}{Mohamed Elhoseiny}, {and} \bibinfo{person}{Leonidas Guibas}.} \bibinfo{year}{2021}\natexlab{}.
\newblock \showarticletitle{ArtEmis: Affective Language for Visual Art}. In \bibinfo{booktitle}{\emph{Proc. IEEE Conf. Comput. Vis. Pattern Recognit.}} \bibinfo{pages}{11564--11574}.
\newblock
\urldef\tempurl%
\url{https://doi.org/10.1109/CVPR46437.2021.01140}
\showDOI{\tempurl}


\bibitem[Antol et~al\mbox{.}(2015)]%
        {VQA}
\bibfield{author}{\bibinfo{person}{Stanislaw Antol}, \bibinfo{person}{Aishwarya Agrawal}, \bibinfo{person}{Jiasen Lu}, \bibinfo{person}{Margaret Mitchell}, \bibinfo{person}{Dhruv Batra}, \bibinfo{person}{C.~Lawrence Zitnick}, {and} \bibinfo{person}{Devi Parikh}.} \bibinfo{year}{2015}\natexlab{}.
\newblock \showarticletitle{{VQA}: Visual Question Answering}. In \bibinfo{booktitle}{\emph{Proc. IEEE Int. Conf. on Comput. Vis.}}
\newblock


\bibitem[Bai et~al\mbox{.}(2023)]%
        {Qwen-VL}
\bibfield{author}{\bibinfo{person}{Jinze Bai}, \bibinfo{person}{Shuai Bai}, \bibinfo{person}{Shusheng Yang}, \bibinfo{person}{Shijie Wang}, \bibinfo{person}{Sinan Tan}, \bibinfo{person}{Peng Wang}, \bibinfo{person}{Junyang Lin}, \bibinfo{person}{Chang Zhou}, {and} \bibinfo{person}{Jingren Zhou}.} \bibinfo{year}{2023}\natexlab{}.
\newblock \showarticletitle{{Qwen-VL}: A Versatile Vision-Language Model for Understanding, Localization, Text Reading, and Beyond}.
\newblock \bibinfo{journal}{\emph{arXiv preprint arXiv:2308.12966}} (\bibinfo{year}{2023}).
\newblock


\bibitem[Chen et~al\mbox{.}(2023b)]%
        {MiniGPT-v2}
\bibfield{author}{\bibinfo{person}{Jun Chen}, \bibinfo{person}{Deyao Zhu}, \bibinfo{person}{Xiaoqian Shen}, \bibinfo{person}{Xiang Li}, \bibinfo{person}{Zechu Liu}, \bibinfo{person}{Pengchuan Zhang}, \bibinfo{person}{Raghuraman Krishnamoorthi}, \bibinfo{person}{Vikas Chandra}, \bibinfo{person}{Yunyang Xiong}, {and} \bibinfo{person}{Mohamed Elhoseiny}.} \bibinfo{year}{2023}\natexlab{b}.
\newblock \showarticletitle{{MiniGPT-v2}: large language model as a unified interface for vision-language multi-task learning}.
\newblock \bibinfo{journal}{\emph{arXiv preprint arXiv:2310.09478}} (\bibinfo{year}{2023}).
\newblock


\bibitem[Chen et~al\mbox{.}(2023a)]%
        {ShareGPT4V}
\bibfield{author}{\bibinfo{person}{Lin Chen}, \bibinfo{person}{Jisong Li}, \bibinfo{person}{Xiaoyi Dong}, \bibinfo{person}{Pan Zhang}, \bibinfo{person}{Conghui He}, \bibinfo{person}{Jiaqi Wang}, \bibinfo{person}{Feng Zhao}, {and} \bibinfo{person}{Dahua Lin}.} \bibinfo{year}{2023}\natexlab{a}.
\newblock \showarticletitle{{ShareGPT4V}: Improving Large Multi-Modal Models with Better Captions}.
\newblock \bibinfo{journal}{\emph{arXiv preprint arXiv:2311.12793}} (\bibinfo{year}{2023}).
\newblock


\bibitem[Chung et~al\mbox{.}(2022)]%
        {Scaling}
\bibfield{author}{\bibinfo{person}{Hyung~Won Chung}, \bibinfo{person}{Le Hou}, \bibinfo{person}{Shayne Longpre}, \bibinfo{person}{Barret Zoph}, \bibinfo{person}{Yi Tay}, \bibinfo{person}{William Fedus}, \bibinfo{person}{Yunxuan Li}, \bibinfo{person}{Xuezhi Wang}, \bibinfo{person}{Mostafa Dehghani}, \bibinfo{person}{Siddhartha Brahma}, \bibinfo{person}{Albert Webson}, \bibinfo{person}{Shixiang~Shane Gu}, \bibinfo{person}{Zhuyun Dai}, \bibinfo{person}{Mirac Suzgun}, \bibinfo{person}{Xinyun Chen}, \bibinfo{person}{Aakanksha Chowdhery}, \bibinfo{person}{Alex Castro-Ros}, \bibinfo{person}{Marie Pellat}, \bibinfo{person}{Kevin Robinson}, \bibinfo{person}{Dasha Valter}, \bibinfo{person}{Sharan Narang}, \bibinfo{person}{Gaurav Mishra}, \bibinfo{person}{Adams Yu}, \bibinfo{person}{Vincent Zhao}, \bibinfo{person}{Yanping Huang}, \bibinfo{person}{Andrew Dai}, \bibinfo{person}{Hongkun Yu}, \bibinfo{person}{Slav Petrov}, \bibinfo{person}{Ed~H. Chi}, \bibinfo{person}{Jeff Dean}, \bibinfo{person}{Jacob Devlin},
  \bibinfo{person}{Adam Roberts}, \bibinfo{person}{Denny Zhou}, \bibinfo{person}{Quoc~V. Le}, {and} \bibinfo{person}{Jason Wei}.} \bibinfo{year}{2022}\natexlab{}.
\newblock \showarticletitle{Scaling Instruction-Finetuned Language Models}.
\newblock \bibinfo{journal}{\emph{arXiv preprint arXiv:2210.11416}} (\bibinfo{year}{2022}).
\newblock


\bibitem[Dai et~al\mbox{.}(2023)]%
        {InstructBLIP}
\bibfield{author}{\bibinfo{person}{Wenliang Dai}, \bibinfo{person}{Junnan Li}, \bibinfo{person}{Dongxu Li}, \bibinfo{person}{Anthony Meng~Huat Tiong}, \bibinfo{person}{Junqi Zhao}, \bibinfo{person}{Weisheng Wang}, \bibinfo{person}{Boyang Li}, \bibinfo{person}{Pascale Fung}, {and} \bibinfo{person}{Steven Hoi}.} \bibinfo{year}{2023}\natexlab{}.
\newblock \showarticletitle{{InstructBLIP}: Towards General-purpose Vision-Language Models with Instruction Tuning}.
\newblock \bibinfo{journal}{\emph{arXiv preprint arXiv:2305.06500}} (\bibinfo{year}{2023}).
\newblock


\bibitem[Deng et~al\mbox{.}(2017)]%
        {AesSurvey}
\bibfield{author}{\bibinfo{person}{Yubin Deng}, \bibinfo{person}{Chen~Change Loy}, {and} \bibinfo{person}{Xiaoou Tang}.} \bibinfo{year}{2017}\natexlab{}.
\newblock \showarticletitle{Image Aesthetic Assessment: An experimental survey}.
\newblock \bibinfo{journal}{\emph{IEEE Signal Process. Mag.}} \bibinfo{volume}{34}, \bibinfo{number}{4} (\bibinfo{year}{2017}), \bibinfo{pages}{80--106}.
\newblock
\urldef\tempurl%
\url{https://doi.org/10.1109/MSP.2017.2696576}
\showDOI{\tempurl}


\bibitem[Du et~al\mbox{.}(2022)]%
        {GLM}
\bibfield{author}{\bibinfo{person}{Zhengxiao Du}, \bibinfo{person}{Yujie Qian}, \bibinfo{person}{Xiao Liu}, \bibinfo{person}{Ming Ding}, \bibinfo{person}{Jiezhong Qiu}, \bibinfo{person}{Zhilin Yang}, {and} \bibinfo{person}{Jie Tang}.} \bibinfo{year}{2022}\natexlab{}.
\newblock \showarticletitle{{GLM}: General Language Model Pretraining with Autoregressive Blank Infilling}.
\newblock \bibinfo{journal}{\emph{arXiv preprint arXiv:2103.10360}} (\bibinfo{year}{2022}).
\newblock


\bibitem[Fang et~al\mbox{.}(2020)]%
        {SPAQ}
\bibfield{author}{\bibinfo{person}{Yuming Fang}, \bibinfo{person}{Hanwei Zhu}, \bibinfo{person}{Yan Zeng}, \bibinfo{person}{Kede Ma}, {and} \bibinfo{person}{Zhou Wang}.} \bibinfo{year}{2020}\natexlab{}.
\newblock \showarticletitle{Perceptual Quality Assessment of Smartphone Photography}. In \bibinfo{booktitle}{\emph{Proc. IEEE Conf. Comput. Vis. Pattern Recognit.}} \bibinfo{pages}{3674--3683}.
\newblock
\urldef\tempurl%
\url{https://doi.org/10.1109/CVPR42600.2020.00373}
\showDOI{\tempurl}


\bibitem[Google(2023)]%
        {Gemini}
\bibfield{author}{\bibinfo{person}{Google}.} \bibinfo{year}{2023}\natexlab{}.
\newblock \bibinfo{booktitle}{\emph{Build with Gemini}}.
\newblock
\urldef\tempurl%
\url{https://ai.google.dev/}
\showURL{%
\tempurl}


\bibitem[Han et~al\mbox{.}(2024)]%
        {Hallucination1}
\bibfield{author}{\bibinfo{person}{Tianyang Han}, \bibinfo{person}{Qing Lian}, \bibinfo{person}{Rui Pan}, \bibinfo{person}{Renjie Pi}, \bibinfo{person}{Jipeng Zhang}, \bibinfo{person}{Shizhe Diao}, \bibinfo{person}{Yong Lin}, {and} \bibinfo{person}{Tong Zhang}.} \bibinfo{year}{2024}\natexlab{}.
\newblock \showarticletitle{The Instinctive Bias: Spurious Images lead to Hallucination in MLLMs}.
\newblock \bibinfo{journal}{\emph{arXiv preprint arXiv:2402.03757}} (\bibinfo{year}{2024}).
\newblock


\bibitem[He et~al\mbox{.}(2022)]%
        {TAD66K}
\bibfield{author}{\bibinfo{person}{Shuai He}, \bibinfo{person}{Yongchang Zhang}, \bibinfo{person}{Rui Xie}, \bibinfo{person}{Dongxiang Jiang}, {and} \bibinfo{person}{Anlong Ming}.} \bibinfo{year}{2022}\natexlab{}.
\newblock \showarticletitle{Rethinking Image Aesthetics Assessment: Models, Datasets and Benchmarks}.
\newblock \bibinfo{journal}{\emph{Proc. Int. Joint Conf. Artif. Intell.}} (\bibinfo{date}{Jul.} \bibinfo{year}{2022}).
\newblock


\bibitem[Hentschel et~al\mbox{.}(2022)]%
        {CLIPAes}
\bibfield{author}{\bibinfo{person}{Simon Hentschel}, \bibinfo{person}{Konstantin Kobs}, {and} \bibinfo{person}{Andreas Hotho}.} \bibinfo{year}{2022}\natexlab{}.
\newblock \showarticletitle{CLIP knows image aesthetics}.
\newblock \bibinfo{journal}{\emph{Frontiers in Artificial Intelligence}}  \bibinfo{volume}{5} (\bibinfo{year}{2022}), \bibinfo{pages}{976235}.
\newblock


\bibitem[Hossain et~al\mbox{.}(2019)]%
        {Captioning}
\bibfield{author}{\bibinfo{person}{MD~Zakir Hossain}, \bibinfo{person}{Ferdous Sohel}, \bibinfo{person}{Mohd~Fairuz Shiratuddin}, {and} \bibinfo{person}{Hamid Laga}.} \bibinfo{year}{2019}\natexlab{}.
\newblock \showarticletitle{A comprehensive survey of deep learning for image captioning}.
\newblock \bibinfo{journal}{\emph{ACM Computing Surveys (CsUR)}} \bibinfo{volume}{51}, \bibinfo{number}{6} (\bibinfo{year}{2019}), \bibinfo{pages}{1--36}.
\newblock


\bibitem[Hosu et~al\mbox{.}(2020)]%
        {KonIQ-10K}
\bibfield{author}{\bibinfo{person}{Vlad Hosu}, \bibinfo{person}{Hanhe Lin}, \bibinfo{person}{Tamas Sziranyi}, {and} \bibinfo{person}{Dietmar Saupe}.} \bibinfo{year}{2020}\natexlab{}.
\newblock \showarticletitle{{KonIQ-10k}: An Ecologically Valid Database for Deep Learning of Blind Image Quality Assessment}.
\newblock \bibinfo{journal}{\emph{IEEE Trans. Image Process.}}  \bibinfo{volume}{29} (\bibinfo{date}{Jan.} \bibinfo{year}{2020}), \bibinfo{pages}{4041--4056}.
\newblock
\urldef\tempurl%
\url{https://doi.org/10.1109/TIP.2020.2967829}
\showDOI{\tempurl}


\bibitem[Huang et~al\mbox{.}(2024b)]%
        {Hallucination2}
\bibfield{author}{\bibinfo{person}{Wen Huang}, \bibinfo{person}{Hongbin Liu}, \bibinfo{person}{Minxin Guo}, {and} \bibinfo{person}{Neil~Zhenqiang Gong}.} \bibinfo{year}{2024}\natexlab{b}.
\newblock \showarticletitle{Visual Hallucinations of Multi-modal Large Language Models}.
\newblock \bibinfo{journal}{\emph{arXiv preprint arXiv:2402.14683}} (\bibinfo{year}{2024}).
\newblock


\bibitem[Huang et~al\mbox{.}(2024a)]%
        {CADAS}
\bibfield{author}{\bibinfo{person}{Yipo Huang}, \bibinfo{person}{Leida Li}, \bibinfo{person}{Pengfei Chen}, \bibinfo{person}{Jinjian Wu}, \bibinfo{person}{Yuzhe Yang}, \bibinfo{person}{Yaqian Li}, {and} \bibinfo{person}{Guangming Shi}.} \bibinfo{year}{2024}\natexlab{a}.
\newblock \showarticletitle{Coarse-to-fine Image Aesthetics Assessment With Dynamic Attribute Selection}.
\newblock \bibinfo{journal}{\emph{IEEE Trans. Multimedia}} (\bibinfo{year}{2024}), \bibinfo{pages}{1--14}.
\newblock


\bibitem[Huang et~al\mbox{.}(2023)]%
        {SARQUE}
\bibfield{author}{\bibinfo{person}{Yipo Huang}, \bibinfo{person}{Leida Li}, \bibinfo{person}{Yuzhe Yang}, \bibinfo{person}{Yaqian Li}, {and} \bibinfo{person}{Yandong Guo}.} \bibinfo{year}{2023}\natexlab{}.
\newblock \showarticletitle{Explainable and Generalizable Blind Image Quality Assessment via Semantic Attribute Reasoning}.
\newblock \bibinfo{journal}{\emph{IEEE Trans. Multimedia}}  \bibinfo{volume}{25} (\bibinfo{year}{2023}), \bibinfo{pages}{7672--7685}.
\newblock
\urldef\tempurl%
\url{https://doi.org/10.1109/TMM.2022.3225728}
\showDOI{\tempurl}


\bibitem[Huang et~al\mbox{.}(2020a)]%
        {IET}
\bibfield{author}{\bibinfo{person}{Yipo Huang}, \bibinfo{person}{Leida Li}, \bibinfo{person}{Yu Zhou}, {and} \bibinfo{person}{Bo Hu}.} \bibinfo{year}{2020}\natexlab{a}.
\newblock \showarticletitle{No-reference quality assessment for live broadcasting videos in temporal and spatial domains}.
\newblock \bibinfo{journal}{\emph{IET Image Processing}} \bibinfo{volume}{14}, \bibinfo{number}{4} (\bibinfo{year}{2020}), \bibinfo{pages}{774--781}.
\newblock


\bibitem[Huang et~al\mbox{.}(2020b)]%
        {DSS}
\bibfield{author}{\bibinfo{person}{Yipo Huang}, \bibinfo{person}{Leida Li}, \bibinfo{person}{Hancheng Zhu}, {and} \bibinfo{person}{Bo Hu}.} \bibinfo{year}{2020}\natexlab{b}.
\newblock \showarticletitle{Blind Quality Index of Depth Images Based on Structural Statistics for View Synthesis}.
\newblock \bibinfo{journal}{\emph{IEEE Signal Process. Lett.}}  \bibinfo{volume}{27} (\bibinfo{date}{Apr.} \bibinfo{year}{2020}), \bibinfo{pages}{685--689}.
\newblock
\urldef\tempurl%
\url{https://doi.org/10.1109/LSP.2020.2988830}
\showDOI{\tempurl}


\bibitem[Huang et~al\mbox{.}(2024c)]%
        {AesBench}
\bibfield{author}{\bibinfo{person}{Yipo Huang}, \bibinfo{person}{Quan Yuan}, \bibinfo{person}{Xiangfei Sheng}, \bibinfo{person}{Zhichao Yang}, \bibinfo{person}{Haoning Wu}, \bibinfo{person}{Pengfei Chen}, \bibinfo{person}{Yuzhe Yang}, \bibinfo{person}{Leida Li}, {and} \bibinfo{person}{Weisi Lin}.} \bibinfo{year}{2024}\natexlab{c}.
\newblock \showarticletitle{AesBench: An Expert Benchmark for Multimodal Large Language Models on Image Aesthetics Perception}.
\newblock \bibinfo{journal}{\emph{arXiv preprint arXiv:2401.08276}} (\bibinfo{year}{2024}).
\newblock


\bibitem[Huggingface(2023)]%
        {IDEFICS}
\bibfield{author}{\bibinfo{person}{Huggingface}.} \bibinfo{year}{2023}\natexlab{}.
\newblock \bibinfo{booktitle}{\emph{Introducing IDEFICS: An open reproduction of state-of-the-art visual language model}}.
\newblock
\urldef\tempurl%
\url{https://huggingface.co/blog/idefics}
\showURL{%
\tempurl}


\bibitem[Hui et~al\mbox{.}(2022a)]%
        {hui2022multi}
\bibfield{author}{\bibinfo{person}{Chen Hui}, \bibinfo{person}{Shaohui Liu}, {and} \bibinfo{person}{Feng Jiang}.} \bibinfo{year}{2022}\natexlab{a}.
\newblock \showarticletitle{Multi-channel adaptive partitioning network for block-based image compressive sensing}. In \bibinfo{booktitle}{\emph{Proc. IEEE Conf. Multimedia Expo,}}. IEEE, \bibinfo{pages}{1--6}.
\newblock


\bibitem[Hui et~al\mbox{.}(2022b)]%
        {hui2022spatio}
\bibfield{author}{\bibinfo{person}{Chen Hui}, \bibinfo{person}{Shaohui Liu}, \bibinfo{person}{Wuzhen Shi}, \bibinfo{person}{Feng Jiang}, {and} \bibinfo{person}{Debin Zhao}.} \bibinfo{year}{2022}\natexlab{b}.
\newblock \showarticletitle{Spatio-temporal context based adaptive camcorder recording watermarking}.
\newblock \bibinfo{journal}{\emph{ACM Trans. Multim. Comput. Commun. Appl.}} \bibinfo{volume}{18}, \bibinfo{number}{3s} (\bibinfo{year}{2022}), \bibinfo{pages}{1--25}.
\newblock


\bibitem[Hui et~al\mbox{.}(2023)]%
        {hui2023rate}
\bibfield{author}{\bibinfo{person}{Chen Hui}, \bibinfo{person}{Shengping Zhang}, \bibinfo{person}{Wenxue Cui}, \bibinfo{person}{Shaohui Liu}, \bibinfo{person}{Feng Jiang}, {and} \bibinfo{person}{Debin Zhao}.} \bibinfo{year}{2023}\natexlab{}.
\newblock \showarticletitle{Rate-adaptive neural network for image compressive sensing}.
\newblock \bibinfo{journal}{\emph{IEEE Trans. Multimedia}} (\bibinfo{year}{2023}).
\newblock


\bibitem[ITU(2012)]%
        {ITU}
\bibfield{author}{\bibinfo{person}{ITU}.} \bibinfo{year}{2012}\natexlab{}.
\newblock \showarticletitle{Methodology for the Subjective Assessment of the Quality of Television Pictures}. In \bibinfo{booktitle}{\emph{Recommendation ITU-R BT.500-13}}. ITU.
\newblock


\bibitem[Ke et~al\mbox{.}(2023)]%
        {VILA}
\bibfield{author}{\bibinfo{person}{Junjie Ke}, \bibinfo{person}{Keren Ye}, \bibinfo{person}{Jiahui Yu}, \bibinfo{person}{Yonghui Wu}, \bibinfo{person}{Peyman Milanfar}, {and} \bibinfo{person}{Feng Yang}.} \bibinfo{year}{2023}\natexlab{}.
\newblock \showarticletitle{VILA: Learning Image Aesthetics from User Comments with Vision-Language Pretraining}. In \bibinfo{booktitle}{\emph{Proc. IEEE Conf. Comput. Vis. Pattern Recog.}} \bibinfo{pages}{10041--10051}.
\newblock
\urldef\tempurl%
\url{https://doi.org/10.1109/CVPR52729.2023.00968}
\showDOI{\tempurl}


\bibitem[Kong et~al\mbox{.}(2016)]%
        {AADB}
\bibfield{author}{\bibinfo{person}{Shu Kong}, \bibinfo{person}{Xiaohui Shen}, \bibinfo{person}{Zhe Lin}, \bibinfo{person}{Radomir Mech}, {and} \bibinfo{person}{Charless Fowlkes}.} \bibinfo{year}{2016}\natexlab{}.
\newblock \showarticletitle{Photo aesthetics ranking network with attributes and content adaptation}. In \bibinfo{booktitle}{\emph{Proc. Eur. Conf. Comput. Vis.}} \bibinfo{pages}{662--679}.
\newblock


\bibitem[Kruk et~al\mbox{.}(2023)]%
        {Impressions}
\bibfield{author}{\bibinfo{person}{Julia Kruk}, \bibinfo{person}{Caleb Ziems}, {and} \bibinfo{person}{Diyi Yang}.} \bibinfo{year}{2023}\natexlab{}.
\newblock \showarticletitle{Impressions: Understanding Visual Semiotics and Aesthetic Impact}.
\newblock \bibinfo{journal}{\emph{arXiv preprint arXiv:2310.17887}} (\bibinfo{year}{2023}).
\newblock


\bibitem[Lai et~al\mbox{.}(2023)]%
        {Segmentation}
\bibfield{author}{\bibinfo{person}{Xin Lai}, \bibinfo{person}{Zhuotao Tian}, \bibinfo{person}{Yukang Chen}, \bibinfo{person}{Yanwei Li}, \bibinfo{person}{Yuhui Yuan}, \bibinfo{person}{Shu Liu}, {and} \bibinfo{person}{Jiaya Jia}.} \bibinfo{year}{2023}\natexlab{}.
\newblock \showarticletitle{{LISA}: Reasoning Segmentation via Large Language Model}.
\newblock \bibinfo{journal}{\emph{arXiv preprint arXiv:2308.00692}} (\bibinfo{year}{2023}).
\newblock


\bibitem[Li et~al\mbox{.}(2018)]%
        {CAD}
\bibfield{author}{\bibinfo{person}{Bo Li}, \bibinfo{person}{Caiming Xiong}, \bibinfo{person}{Tianfu Wu}, \bibinfo{person}{Yu Zhou}, \bibinfo{person}{Lun Zhang}, {and} \bibinfo{person}{Rufeng Chu}.} \bibinfo{year}{2018}\natexlab{}.
\newblock \showarticletitle{Neural Abstract Style Transfer for Chinese Traditional Painting}.
\newblock \bibinfo{journal}{\emph{arXiv preprint arXiv:1812.03264}} (\bibinfo{year}{2018}).
\newblock


\bibitem[Li et~al\mbox{.}(2023b)]%
        {Otter}
\bibfield{author}{\bibinfo{person}{Bo Li}, \bibinfo{person}{Yuanhan Zhang}, \bibinfo{person}{Liangyu Chen}, \bibinfo{person}{Jinghao Wang}, \bibinfo{person}{Jingkang Yang}, {and} \bibinfo{person}{Ziwei Liu}.} \bibinfo{year}{2023}\natexlab{b}.
\newblock \showarticletitle{Otter: A Multi-Modal Model with In-Context Instruction Tuning}.
\newblock \bibinfo{journal}{\emph{arXiv preprint arXiv:2305.03726}} (\bibinfo{year}{2023}).
\newblock


\bibitem[Li et~al\mbox{.}(2023c)]%
        {AGIQA-3K}
\bibfield{author}{\bibinfo{person}{Chunyi Li}, \bibinfo{person}{Zicheng Zhang}, \bibinfo{person}{Haoning Wu}, \bibinfo{person}{Wei Sun}, \bibinfo{person}{Xiongkuo Min}, \bibinfo{person}{Xiaohong Liu}, \bibinfo{person}{Guangtao Zhai}, {and} \bibinfo{person}{Weisi Lin}.} \bibinfo{year}{2023}\natexlab{c}.
\newblock \showarticletitle{{AGIQA-3K}: An Open Database for AI-Generated Image Quality Assessment}.
\newblock \bibinfo{journal}{\emph{arXiv preprint arXiv:2306.04717}} (\bibinfo{year}{2023}).
\newblock


\bibitem[Li et~al\mbox{.}(2023a)]%
        {TAVAR}
\bibfield{author}{\bibinfo{person}{Leida Li}, \bibinfo{person}{Yipo Huang}, \bibinfo{person}{Jinjian Wu}, \bibinfo{person}{Yuzhe Yang}, \bibinfo{person}{Yaqian Li}, \bibinfo{person}{Yandong Guo}, {and} \bibinfo{person}{Guangming Shi}.} \bibinfo{year}{2023}\natexlab{a}.
\newblock \showarticletitle{Theme-aware Visual Attribute Reasoning for Image Aesthetics Assessment}.
\newblock \bibinfo{journal}{\emph{IEEE Trans. Circuits and Syst. Video Technol.}} (\bibinfo{year}{2023}).
\newblock
\urldef\tempurl%
\url{https://doi.org/10.1109/TCSVT.2023.3249185}
\showDOI{\tempurl}


\bibitem[Li et~al\mbox{.}(2020)]%
        {Zhu-TIP}
\bibfield{author}{\bibinfo{person}{Leida Li}, \bibinfo{person}{Hancheng Zhu}, \bibinfo{person}{Sicheng Zhao}, \bibinfo{person}{Guiguang Ding}, {and} \bibinfo{person}{Weisi Lin}.} \bibinfo{year}{2020}\natexlab{}.
\newblock \showarticletitle{Personality-assisted multi-task learning for generic and personalized image aesthetics assessment}.
\newblock \bibinfo{journal}{\emph{IEEE Trans. Image Process.}}  \bibinfo{volume}{29} (\bibinfo{date}{Jan.} \bibinfo{year}{2020}), \bibinfo{pages}{3898--3910}.
\newblock


\bibitem[Lin et~al\mbox{.}(2015)]%
        {COCO}
\bibfield{author}{\bibinfo{person}{Tsung-Yi Lin}, \bibinfo{person}{Michael Maire}, \bibinfo{person}{Serge Belongie}, \bibinfo{person}{Lubomir Bourdev}, \bibinfo{person}{Ross Girshick}, \bibinfo{person}{James Hays}, \bibinfo{person}{Pietro Perona}, \bibinfo{person}{Deva Ramanan}, \bibinfo{person}{C.~Lawrence Zitnick}, {and} \bibinfo{person}{Piotr Dollár}.} \bibinfo{year}{2015}\natexlab{}.
\newblock \showarticletitle{Microsoft COCO: Common Objects in Context}.
\newblock \bibinfo{journal}{\emph{arXiv preprint arXiv:1405.0312}} (\bibinfo{year}{2015}).
\newblock


\bibitem[Liu et~al\mbox{.}(2023a)]%
        {LLaVA-1.5}
\bibfield{author}{\bibinfo{person}{Haotian Liu}, \bibinfo{person}{Chunyuan Li}, \bibinfo{person}{Yuheng Li}, {and} \bibinfo{person}{Yong~Jae Lee}.} \bibinfo{year}{2023}\natexlab{a}.
\newblock \showarticletitle{Improved Baselines with Visual Instruction Tuning}.
\newblock \bibinfo{journal}{\emph{arXiv preprint arXiv:2310.03744}} (\bibinfo{year}{2023}).
\newblock


\bibitem[Liu et~al\mbox{.}(2023b)]%
        {LLAVA}
\bibfield{author}{\bibinfo{person}{Haotian Liu}, \bibinfo{person}{Chunyuan Li}, \bibinfo{person}{Qingyang Wu}, {and} \bibinfo{person}{Yong~Jae Lee}.} \bibinfo{year}{2023}\natexlab{b}.
\newblock \showarticletitle{Visual Instruction Tuning}.
\newblock \bibinfo{journal}{\emph{arXiv preprint arXiv:2304.08485}} (\bibinfo{year}{2023}).
\newblock


\bibitem[Lu et~al\mbox{.}(2022)]%
        {ScienceQA}
\bibfield{author}{\bibinfo{person}{Pan Lu}, \bibinfo{person}{Swaroop Mishra}, \bibinfo{person}{Tony Xia}, \bibinfo{person}{Liang Qiu}, \bibinfo{person}{Kai-Wei Chang}, \bibinfo{person}{Song-Chun Zhu}, \bibinfo{person}{Oyvind Tafjord}, \bibinfo{person}{Peter Clark}, {and} \bibinfo{person}{Ashwin Kalyan}.} \bibinfo{year}{2022}\natexlab{}.
\newblock \showarticletitle{Learn to Explain: Multimodal Reasoning via Thought Chains for Science Question Answering}. In \bibinfo{booktitle}{\emph{Proc. Neural Inf. Process. Syst.}} \bibinfo{pages}{2507--2521}.
\newblock


\bibitem[Mi et~al\mbox{.}(2023a)]%
        {mi2023clif}
\bibfield{author}{\bibinfo{person}{Yachun Mi}, \bibinfo{person}{Yu Li}, \bibinfo{person}{Yan Shu}, \bibinfo{person}{Chen Hui}, \bibinfo{person}{Puchao Zhou}, {and} \bibinfo{person}{Shaohui Liu}.} \bibinfo{year}{2023}\natexlab{a}.
\newblock \showarticletitle{CLiF-VQA: Enhancing Video Quality Assessment by Incorporating High-Level Semantic Information related to Human Feelings}.
\newblock \bibinfo{journal}{\emph{arXiv preprint arXiv:2311.07090}} (\bibinfo{year}{2023}).
\newblock


\bibitem[Mi et~al\mbox{.}(2024)]%
        {mi2024ze}
\bibfield{author}{\bibinfo{person}{Yachun Mi}, \bibinfo{person}{Yu Li}, \bibinfo{person}{Yan Shu}, {and} \bibinfo{person}{Shaohui Liu}.} \bibinfo{year}{2024}\natexlab{}.
\newblock \showarticletitle{ZE-FESG: A Zero-Shot Feature Extraction Method Based on Semantic Guidance for No-Reference Video Quality Assessment}. In \bibinfo{booktitle}{\emph{Proc. IEEE Int. Conf. Acoust. Speech Signal Process.}} IEEE, \bibinfo{pages}{3640--3644}.
\newblock


\bibitem[Mi et~al\mbox{.}(2023b)]%
        {mi2023vva}
\bibfield{author}{\bibinfo{person}{Yachun Mi}, \bibinfo{person}{Yan Shu}, \bibinfo{person}{Honglei Xu}, \bibinfo{person}{Shaohui Liu}, {and} \bibinfo{person}{Feng Jiang}.} \bibinfo{year}{2023}\natexlab{b}.
\newblock \showarticletitle{VVA: Video Values Analysis}. In \bibinfo{booktitle}{\emph{Chinese Conf. Pattern Recog. Comput. Vis.}} Springer, \bibinfo{pages}{346--358}.
\newblock


\bibitem[Mishra et~al\mbox{.}(2019)]%
        {OCR-VQA}
\bibfield{author}{\bibinfo{person}{Anand Mishra}, \bibinfo{person}{Shashank Shekhar}, \bibinfo{person}{Ajeet~Kumar Singh}, {and} \bibinfo{person}{Anirban Chakraborty}.} \bibinfo{year}{2019}\natexlab{}.
\newblock \showarticletitle{{OCR-VQA}: Visual Question Answering by Reading Text in Images}. In \bibinfo{booktitle}{\emph{2019 International Conference on Document Analysis and Recognition (ICDAR)}}. \bibinfo{pages}{947--952}.
\newblock
\urldef\tempurl%
\url{https://doi.org/10.1109/ICDAR.2019.00156}
\showDOI{\tempurl}


\bibitem[Murray et~al\mbox{.}(2012)]%
        {AVA}
\bibfield{author}{\bibinfo{person}{Naila Murray}, \bibinfo{person}{Luca Marchesotti}, {and} \bibinfo{person}{Florent Perronnin}.} \bibinfo{year}{2012}\natexlab{}.
\newblock \showarticletitle{{AVA}: A large-scale database for aesthetic visual analysis}. In \bibinfo{booktitle}{\emph{Proc. IEEE Conf. Comput. Vis. Pattern Recog.}} \bibinfo{pages}{2408--2415}.
\newblock


\bibitem[OpenAI(2023)]%
        {gpt4}
\bibfield{author}{\bibinfo{person}{OpenAI}.} \bibinfo{year}{2023}\natexlab{}.
\newblock \showarticletitle{{GPT-4} Technical Report}.
\newblock \bibinfo{journal}{\emph{arXiv preprint arXiv:2303.08774}} (\bibinfo{year}{2023}).
\newblock


\bibitem[Radford et~al\mbox{.}(2021)]%
        {CLIP}
\bibfield{author}{\bibinfo{person}{Alec Radford}, \bibinfo{person}{Jong~Wook Kim}, \bibinfo{person}{Chris Hallacy}, \bibinfo{person}{Aditya Ramesh}, \bibinfo{person}{Gabriel Goh}, \bibinfo{person}{Sandhini Agarwal}, \bibinfo{person}{Girish Sastry}, \bibinfo{person}{Amanda Askell}, \bibinfo{person}{Pamela Mishkin}, \bibinfo{person}{Jack Clark}, \bibinfo{person}{Gretchen Krueger}, {and} \bibinfo{person}{Ilya Sutskever}.} \bibinfo{year}{2021}\natexlab{}.
\newblock \showarticletitle{Learning Transferable Visual Models From Natural Language Supervision}.
\newblock \bibinfo{journal}{\emph{arXiv preprint arXiv:2103.00020}} (\bibinfo{year}{2021}).
\newblock


\bibitem[Sheng et~al\mbox{.}(2023)]%
        {AesCLIP}
\bibfield{author}{\bibinfo{person}{Xiangfei Sheng}, \bibinfo{person}{Leida Li}, \bibinfo{person}{Pengfei Chen}, \bibinfo{person}{Jinjian Wu}, \bibinfo{person}{Weisheng Dong}, \bibinfo{person}{Yuzhe Yang}, \bibinfo{person}{Liwu Xu}, \bibinfo{person}{Yaqian Li}, {and} \bibinfo{person}{Guangming Shi}.} \bibinfo{year}{2023}\natexlab{}.
\newblock \showarticletitle{AesCLIP: Multi-Attribute Contrastive Learning for Image Aesthetics Assessment}. In \bibinfo{booktitle}{\emph{Proc. ACM Int. Conf. Multimedia}}. \bibinfo{pages}{1117–1126}.
\newblock


\bibitem[Touvron et~al\mbox{.}(2023)]%
        {LLaMA}
\bibfield{author}{\bibinfo{person}{Hugo Touvron}, \bibinfo{person}{Thibaut Lavril}, \bibinfo{person}{Gautier Izacard}, \bibinfo{person}{Xavier Martinet}, \bibinfo{person}{Marie-Anne Lachaux}, \bibinfo{person}{Timothée Lacroix}, \bibinfo{person}{Baptiste Rozière}, \bibinfo{person}{Naman Goyal}, \bibinfo{person}{Eric Hambro}, \bibinfo{person}{Faisal Azhar}, \bibinfo{person}{Aurelien Rodriguez}, \bibinfo{person}{Armand Joulin}, \bibinfo{person}{Edouard Grave}, {and} \bibinfo{person}{Guillaume Lample}.} \bibinfo{year}{2023}\natexlab{}.
\newblock \showarticletitle{{LLaMA}: Open and Efficient Foundation Language Models}.
\newblock \bibinfo{journal}{\emph{arXiv preprint arXiv:2302.13971}} (\bibinfo{year}{2023}).
\newblock


\bibitem[Unsplash(2023)]%
        {LITE}
\bibfield{author}{\bibinfo{person}{Unsplash}.} \bibinfo{year}{2023}\natexlab{}.
\newblock \bibinfo{booktitle}{\emph{Access the world’s largest open library dataset}}.
\newblock
\urldef\tempurl%
\url{https://unsplash.com/data}
\showURL{%
\tempurl}


\bibitem[Wang et~al\mbox{.}(2023)]%
        {DiffusionDB}
\bibfield{author}{\bibinfo{person}{Zijie~J. Wang}, \bibinfo{person}{Evan Montoya}, \bibinfo{person}{David Munechika}, \bibinfo{person}{Haoyang Yang}, \bibinfo{person}{Benjamin Hoover}, {and} \bibinfo{person}{Duen~Horng Chau}.} \bibinfo{year}{2023}\natexlab{}.
\newblock \showarticletitle{{DiffusionDB}: A Large-scale Prompt Gallery Dataset for Text-to-Image Generative Models}.
\newblock \bibinfo{journal}{\emph{arXiv preprint arXiv:2210.14896}} (\bibinfo{year}{2023}).
\newblock


\bibitem[Wu et~al\mbox{.}(2024a)]%
        {Q-Bench}
\bibfield{author}{\bibinfo{person}{Haoning Wu}, \bibinfo{person}{Zicheng Zhang}, \bibinfo{person}{Erli Zhang}, \bibinfo{person}{Chaofeng Chen}, \bibinfo{person}{Liang Liao}, \bibinfo{person}{Annan Wang}, \bibinfo{person}{Chunyi Li}, \bibinfo{person}{Wenxiu Sun}, \bibinfo{person}{Qiong Yan}, \bibinfo{person}{Guangtao Zhai}, {and} \bibinfo{person}{Weisi Lin}.} \bibinfo{year}{2024}\natexlab{a}.
\newblock \showarticletitle{{Q-Bench}: A Benchmark for General-Purpose Foundation Models on Low-level Vision}.
\newblock \bibinfo{journal}{\emph{arXiv preprint arXiv:2309.14181}} (\bibinfo{year}{2024}).
\newblock


\bibitem[Wu et~al\mbox{.}(2023b)]%
        {Q-Instruct}
\bibfield{author}{\bibinfo{person}{Haoning Wu}, \bibinfo{person}{Zicheng Zhang}, \bibinfo{person}{Erli Zhang}, \bibinfo{person}{Chaofeng Chen}, \bibinfo{person}{Liang Liao}, \bibinfo{person}{Annan Wang}, \bibinfo{person}{Kaixin Xu}, \bibinfo{person}{Chunyi Li}, \bibinfo{person}{Jingwen Hou}, \bibinfo{person}{Guangtao Zhai}, \bibinfo{person}{Geng Xue}, \bibinfo{person}{Wenxiu Sun}, \bibinfo{person}{Qiong Yan}, {and} \bibinfo{person}{Weisi Lin}.} \bibinfo{year}{2023}\natexlab{b}.
\newblock \showarticletitle{{Q-Instruct}: Improving Low-level Visual Abilities for Multi-modality Foundation Models}.
\newblock \bibinfo{journal}{\emph{arXiv preprint arXiv:2311.06783}} (\bibinfo{year}{2023}).
\newblock


\bibitem[Wu et~al\mbox{.}(2023a)]%
        {Q-Align}
\bibfield{author}{\bibinfo{person}{Haoning Wu}, \bibinfo{person}{Zicheng Zhang}, \bibinfo{person}{Weixia Zhang}, \bibinfo{person}{Chaofeng Chen}, \bibinfo{person}{Chunyi Li}, \bibinfo{person}{Liang Liao}, \bibinfo{person}{Annan Wang}, \bibinfo{person}{Erli Zhang}, \bibinfo{person}{Wenxiu Sun}, \bibinfo{person}{Qiong Yan}, \bibinfo{person}{Xiongkuo Min}, \bibinfo{person}{Guangtai Zhai}, {and} \bibinfo{person}{Weisi Lin}.} \bibinfo{year}{2023}\natexlab{a}.
\newblock \showarticletitle{Q-Align: Teaching LMMs for Visual Scoring via Discrete Text-Defined Levels}.
\newblock \bibinfo{journal}{\emph{arXiv preprint arXiv:2312.17090}} (\bibinfo{year}{2023}).
\newblock


\bibitem[Wu et~al\mbox{.}(2024b)]%
        {co-instruct}
\bibfield{author}{\bibinfo{person}{Haoning Wu}, \bibinfo{person}{Hanwei Zhu}, \bibinfo{person}{Zicheng Zhang}, \bibinfo{person}{Erli Zhang}, \bibinfo{person}{Chaofeng Chen}, \bibinfo{person}{Liang Liao}, \bibinfo{person}{Chunyi Li}, \bibinfo{person}{Annan Wang}, \bibinfo{person}{Wenxiu Sun}, \bibinfo{person}{Qiong Yan}, \bibinfo{person}{Xiaohong Liu}, \bibinfo{person}{Guangtao Zhai}, \bibinfo{person}{Shiqi Wang}, {and} \bibinfo{person}{Weisi Lin}.} \bibinfo{year}{2024}\natexlab{b}.
\newblock \bibinfo{title}{Towards Open-ended Visual Quality Comparison}.
\newblock
\newblock


\bibitem[Xu et~al\mbox{.}(2012b)]%
        {xu-5}
\bibfield{author}{\bibinfo{person}{Qianqian Xu}, \bibinfo{person}{Qingming Huang}, \bibinfo{person}{Tingting Jiang}, \bibinfo{person}{Bowei Yan}, \bibinfo{person}{Weisi Lin}, {and} \bibinfo{person}{Yuan Yao}.} \bibinfo{year}{2012}\natexlab{b}.
\newblock \showarticletitle{HodgeRank on Random Graphs for Subjective Video Quality Assessment}.
\newblock \bibinfo{journal}{\emph{IEEE Trans. Multimedia}} \bibinfo{volume}{14}, \bibinfo{number}{3} (\bibinfo{year}{2012}), \bibinfo{pages}{844--857}.
\newblock


\bibitem[Xu et~al\mbox{.}(2012a)]%
        {xu-4}
\bibfield{author}{\bibinfo{person}{Qianqian Xu}, \bibinfo{person}{Qingming Huang}, {and} \bibinfo{person}{Yuan Yao}.} \bibinfo{year}{2012}\natexlab{a}.
\newblock \showarticletitle{Online crowdsourcing subjective image quality assessment}. In \bibinfo{booktitle}{\emph{Proc. ACM Int. Conf. Multimedia}}. \bibinfo{pages}{359--368}.
\newblock


\bibitem[Xu et~al\mbox{.}(2011)]%
        {xu-6}
\bibfield{author}{\bibinfo{person}{Qianqian Xu}, \bibinfo{person}{Tingting Jiang}, \bibinfo{person}{Yuan Yao}, \bibinfo{person}{Qingming Huang}, \bibinfo{person}{Bowei Yan}, {and} \bibinfo{person}{Weisi Lin}.} \bibinfo{year}{2011}\natexlab{}.
\newblock \showarticletitle{Random partial paired comparison for subjective video quality assessment via HodgeRank}. In \bibinfo{booktitle}{\emph{Proc. ACM Int. Conf. Multimedia}}. \bibinfo{pages}{393--402}.
\newblock


\bibitem[Xu et~al\mbox{.}(2013)]%
        {xu-3}
\bibfield{author}{\bibinfo{person}{Qianqian Xu}, \bibinfo{person}{Jiechao Xiong}, \bibinfo{person}{Qingming Huang}, {and} \bibinfo{person}{Yuan Yao}.} \bibinfo{year}{2013}\natexlab{}.
\newblock \showarticletitle{Robust evaluation for quality of experience in crowdsourcing}. In \bibinfo{booktitle}{\emph{Proc. ACM Int. Conf. Multimedia}}. \bibinfo{pages}{43--52}.
\newblock


\bibitem[Yang et~al\mbox{.}(2022)]%
        {PARA}
\bibfield{author}{\bibinfo{person}{Yuzhe Yang}, \bibinfo{person}{Liwu Xu}, \bibinfo{person}{Leida Li}, \bibinfo{person}{Nan Qie}, \bibinfo{person}{Yaqian Li}, \bibinfo{person}{Peng Zhang}, {and} \bibinfo{person}{Yandong Guo}.} \bibinfo{year}{2022}\natexlab{}.
\newblock \showarticletitle{Personalized Image Aesthetics Assessment with Rich Attributes}. In \bibinfo{booktitle}{\emph{Proc. IEEE Conf. Comput. Vis. Pattern Recog.}} \bibinfo{pages}{19861--19869}.
\newblock


\bibitem[Yang et~al\mbox{.}(2023)]%
        {GPT-4V}
\bibfield{author}{\bibinfo{person}{Zhengyuan Yang}, \bibinfo{person}{Linjie Li}, \bibinfo{person}{Kevin Lin}, \bibinfo{person}{Jianfeng Wang}, \bibinfo{person}{Chung-Ching Lin}, \bibinfo{person}{Zicheng Liu}, {and} \bibinfo{person}{Lijuan Wang}.} \bibinfo{year}{2023}\natexlab{}.
\newblock \showarticletitle{The Dawn of LMMs: Preliminary Explorations with GPT-4V(ision)}.
\newblock \bibinfo{journal}{\emph{arXiv preprint arXiv:2309.17421}} (\bibinfo{year}{2023}).
\newblock


\bibitem[Yang et~al\mbox{.}(2024)]%
        {Zhichao}
\bibfield{author}{\bibinfo{person}{Zhichao Yang}, \bibinfo{person}{Leida Li}, \bibinfo{person}{Yuzhe Yang}, \bibinfo{person}{Yaqian Li}, {and} \bibinfo{person}{Weisi Lin}.} \bibinfo{year}{2024}\natexlab{}.
\newblock \showarticletitle{Multi-Level Transitional Contrast Learning for Personalized Image Aesthetics Assessment}.
\newblock \bibinfo{journal}{\emph{IEEE Trans. Multimedia}}  \bibinfo{volume}{26} (\bibinfo{year}{2024}), \bibinfo{pages}{1944--1956}.
\newblock


\bibitem[Ye et~al\mbox{.}(2023a)]%
        {mPLUG-Owl}
\bibfield{author}{\bibinfo{person}{Qinghao Ye}, \bibinfo{person}{Haiyang Xu}, \bibinfo{person}{Guohai Xu}, \bibinfo{person}{Jiabo Ye}, \bibinfo{person}{Ming Yan}, \bibinfo{person}{Yiyang Zhou}, \bibinfo{person}{Junyang Wang}, \bibinfo{person}{Anwen Hu}, \bibinfo{person}{Pengcheng Shi}, \bibinfo{person}{Yaya Shi}, \bibinfo{person}{Chenliang Li}, \bibinfo{person}{Yuanhong Xu}, \bibinfo{person}{Hehong Chen}, \bibinfo{person}{Junfeng Tian}, \bibinfo{person}{Qian Qi}, \bibinfo{person}{Ji Zhang}, {and} \bibinfo{person}{Fei Huang}.} \bibinfo{year}{2023}\natexlab{a}.
\newblock \showarticletitle{{mPLUG-Owl}: Modularization Empowers Large Language Models with Multimodality}.
\newblock \bibinfo{journal}{\emph{arXiv preprint arXiv:2304.14178}} (\bibinfo{year}{2023}).
\newblock


\bibitem[Ye et~al\mbox{.}(2023b)]%
        {mPLUG-Owl2}
\bibfield{author}{\bibinfo{person}{Qinghao Ye}, \bibinfo{person}{Haiyang Xu}, \bibinfo{person}{Jiabo Ye}, \bibinfo{person}{Ming Yan}, \bibinfo{person}{Anwen Hu}, \bibinfo{person}{Haowei Liu}, \bibinfo{person}{Qi Qian}, \bibinfo{person}{Ji Zhang}, \bibinfo{person}{Fei Huang}, {and} \bibinfo{person}{Jingren Zhou}.} \bibinfo{year}{2023}\natexlab{b}.
\newblock \showarticletitle{{mPLUG-Owl2}: Revolutionizing Multi-modal Large Language Model with Modality Collaboration}.
\newblock \bibinfo{journal}{\emph{arXiv preprint arXiv:2311.04257}} (\bibinfo{year}{2023}).
\newblock


\bibitem[Yi et~al\mbox{.}(2023)]%
        {BAID}
\bibfield{author}{\bibinfo{person}{Ran Yi}, \bibinfo{person}{Haoyuan Tian}, \bibinfo{person}{Zhihao Gu}, \bibinfo{person}{Yu-Kun Lai}, {and} \bibinfo{person}{Paul~L. Rosin}.} \bibinfo{year}{2023}\natexlab{}.
\newblock \showarticletitle{Towards Artistic Image Aesthetics Assessment: a Large-scale Dataset and a New Method}. In \bibinfo{booktitle}{\emph{Proc. IEEE Conf. Comput. Vis. Pattern Recognit.}} \bibinfo{pages}{22388--22397}.
\newblock
\urldef\tempurl%
\url{https://doi.org/10.1109/CVPR52729.2023.02144}
\showDOI{\tempurl}


\bibitem[Yuan et~al\mbox{.}(2023)]%
        {TinyGPT-V}
\bibfield{author}{\bibinfo{person}{Zhengqing Yuan}, \bibinfo{person}{Zhaoxu Li}, {and} \bibinfo{person}{Lichao Sun}.} \bibinfo{year}{2023}\natexlab{}.
\newblock \showarticletitle{{TinyGPT-V}: Efficient Multimodal Large Language Model via Small Backbones}.
\newblock \bibinfo{journal}{\emph{arXiv preprint arXiv:2312.16862}} (\bibinfo{year}{2023}).
\newblock


\bibitem[Zhang et~al\mbox{.}(2024)]%
        {Survey}
\bibfield{author}{\bibinfo{person}{Shengyu Zhang}, \bibinfo{person}{Linfeng Dong}, \bibinfo{person}{Xiaoya Li}, \bibinfo{person}{Sen Zhang}, \bibinfo{person}{Xiaofei Sun}, \bibinfo{person}{Shuhe Wang}, \bibinfo{person}{Jiwei Li}, \bibinfo{person}{Runyi Hu}, \bibinfo{person}{Tianwei Zhang}, \bibinfo{person}{Fei Wu}, {and} \bibinfo{person}{Guoyin Wang}.} \bibinfo{year}{2024}\natexlab{}.
\newblock \showarticletitle{Instruction Tuning for Large Language Models: A Survey}.
\newblock \bibinfo{journal}{\emph{arXiv preprint arXiv:2308.10792}} (\bibinfo{year}{2024}).
\newblock


\bibitem[Zhang et~al\mbox{.}(2023)]%
        {AGIQA-1K}
\bibfield{author}{\bibinfo{person}{Zicheng Zhang}, \bibinfo{person}{Chunyi Li}, \bibinfo{person}{Wei Sun}, \bibinfo{person}{Xiaohong Liu}, \bibinfo{person}{Xiongkuo Min}, {and} \bibinfo{person}{Guangtao Zhai}.} \bibinfo{year}{2023}\natexlab{}.
\newblock \showarticletitle{A Perceptual Quality Assessment Exploration for AIGC Images}.
\newblock \bibinfo{journal}{\emph{arXiv preprint arXiv:2303.12618}} (\bibinfo{year}{2023}).
\newblock


\bibitem[Zheng et~al\mbox{.}(2024)]%
        {Judging}
\bibfield{author}{\bibinfo{person}{Lianmin Zheng}, \bibinfo{person}{Wei-Lin Chiang}, \bibinfo{person}{Ying Sheng}, \bibinfo{person}{Siyuan Zhuang}, \bibinfo{person}{Zhanghao Wu}, \bibinfo{person}{Yonghao Zhuang}, \bibinfo{person}{Zi Lin}, \bibinfo{person}{Zhuohan Li}, \bibinfo{person}{Dacheng Li}, \bibinfo{person}{Eric Xing}, {et~al\mbox{.}}} \bibinfo{year}{2024}\natexlab{}.
\newblock \showarticletitle{Judging llm-as-a-judge with mt-bench and chatbot arena}.
\newblock \bibinfo{journal}{\emph{Advances in Neural Information Processing Systems}}  \bibinfo{volume}{36} (\bibinfo{year}{2024}).
\newblock


\bibitem[Zhu et~al\mbox{.}(2023a)]%
        {MiniGPT-4}
\bibfield{author}{\bibinfo{person}{Deyao Zhu}, \bibinfo{person}{Jun Chen}, \bibinfo{person}{Xiaoqian Shen}, \bibinfo{person}{Xiang Li}, {and} \bibinfo{person}{Mohamed Elhoseiny}.} \bibinfo{year}{2023}\natexlab{a}.
\newblock \showarticletitle{{MiniGPT-4}: Enhancing Vision-Language Understanding with Advanced Large Language Models}.
\newblock \bibinfo{journal}{\emph{arXiv preprint arXiv:2304.10592}} (\bibinfo{year}{2023}).
\newblock


\bibitem[Zhu et~al\mbox{.}(2022)]%
        {Zhu3}
\bibfield{author}{\bibinfo{person}{Hancheng Zhu}, \bibinfo{person}{Leida Li}, \bibinfo{person}{Jinjian Wu}, \bibinfo{person}{Sicheng Zhao}, \bibinfo{person}{Guiguang Ding}, {and} \bibinfo{person}{Guangming Shi}.} \bibinfo{year}{2022}\natexlab{}.
\newblock \showarticletitle{Personalized Image Aesthetics Assessment via Meta-Learning With Bilevel Gradient Optimization}.
\newblock \bibinfo{journal}{\emph{IEEE Trans. Cybern.}} \bibinfo{volume}{52}, \bibinfo{number}{3} (\bibinfo{year}{2022}), \bibinfo{pages}{1798--1811}.
\newblock


\bibitem[Zhu et~al\mbox{.}(2023b)]%
        {zhu2023personalized}
\bibfield{author}{\bibinfo{person}{Hancheng Zhu}, \bibinfo{person}{Zhiwen Shao}, \bibinfo{person}{Yong Zhou}, \bibinfo{person}{Guangcheng Wang}, \bibinfo{person}{Pengfei Chen}, {and} \bibinfo{person}{Leida Li}.} \bibinfo{year}{2023}\natexlab{b}.
\newblock \showarticletitle{Personalized Image Aesthetics Assessment with Attribute-guided Fine-grained Feature Representation}. In \bibinfo{booktitle}{\emph{Proc. ACM Int. Conf. Multimedia}}. \bibinfo{pages}{6794--6802}.
\newblock


\bibitem[Zhu et~al\mbox{.}(2023c)]%
        {Zhu2}
\bibfield{author}{\bibinfo{person}{Hancheng Zhu}, \bibinfo{person}{Yong Zhou}, \bibinfo{person}{Leida Li}, \bibinfo{person}{Yaqian Li}, {and} \bibinfo{person}{Yandong Guo}.} \bibinfo{year}{2023}\natexlab{c}.
\newblock \showarticletitle{Learning Personalized Image Aesthetics From Subjective and Objective Attributes}.
\newblock \bibinfo{journal}{\emph{IEEE Trans. Multimedia}}  \bibinfo{volume}{25} (\bibinfo{year}{2023}), \bibinfo{pages}{179--190}.
\newblock


\end{thebibliography}










\end{document}